\documentclass{article}

\usepackage{microtype}
\usepackage{graphicx}
\usepackage{subfigure}
\usepackage{booktabs}
\usepackage{multirow}

\usepackage{hyperref}

\usepackage{longtable}
\usepackage{placeins}
\usepackage{array}
\usepackage{enumitem}
\usepackage{url}
\usepackage{amsmath}
\usepackage{amssymb}
\usepackage{amsthm}
\usepackage{mathtools}
\usepackage{newtxtext,newtxmath}

\usepackage{pgfplots}
\pgfplotsset{compat=1.18}
\usepgfplotslibrary{groupplots,fillbetween}
\usetikzlibrary{calc,plotmarks,arrows.meta}

\usepackage[colorinlistoftodos,prependcaption,textsize=tiny]{todonotes}
\definecolor{yasongtodored}{HTML}{B2182B}
\NewDocumentCommand{\yasongtodo}{O{} m}{%
  \todo[
    inline,
    linecolor=yasongtodored,
    backgroundcolor=yasongtodored!15,
    bordercolor=yasongtodored,
    #1
  ]{Yasong: #2}%
}
\NewDocumentCommand{\yasongmargintodo}{O{} m}{%
  \begingroup
  \normalmarginpar
  \todo[
    linecolor=yasongtodored,
    backgroundcolor=yasongtodored!15,
    bordercolor=yasongtodored,
    #1
  ]{Yasong: #2}%
  \endgroup
}

\definecolor{sivatodoblue}{HTML}{2166AC}
\NewDocumentCommand{\sivatodo}{O{} m}{%
  \todo[
    inline,
    linecolor=sivatodoblue,
    backgroundcolor=sivatodoblue!15,
    bordercolor=sivatodoblue,
    #1
  ]{Siva: #2}%
}
\NewDocumentCommand{\sivamargintodo}{O{} m}{%
  \begingroup
  \normalmarginpar
  \todo[
    linecolor=sivatodoblue,
    backgroundcolor=sivatodoblue!15,
    bordercolor=sivatodoblue,
    #1
  ]{Siva: #2}%
  \endgroup
}

\newtheorem{theorem}{Theorem}
\newtheorem{lemma}{Lemma}

\newtheorem{assumption}{Assumption}
\theoremstyle{remark}

\newcommand{\cX}{\mathcal{X}}

\newcommand{\E}{\mathbb{E}}

\newcommand{\mask}{\mathtt{mask}}
\newcommand{\pospart}[1]{\left[#1\right]_{+}}

\def\method{\textsc{SpecCTMC} }

\usepackage{hyperref}
\usepackage[capitalize,noabbrev]{cleveref}
\crefname{assumption}{assumption}{assumptions}
\Crefname{assumption}{Assumption}{Assumptions}

\makeatletter
\let\standardaddcontentsline\addcontentsline
\makeatother

\usepackage[accepted]{mlsys2025}

\makeatletter
\renewcommand{\Notice@String}{}
\makeatother
\hypersetup{pdfsubject={}}

\mlsystitlerunning{Exact Speculative CTMC Simulation}

\begin{document}

\twocolumn[
\mlsystitle{\method: Exact Speculative Simulation of Diffusion Language Model Continuous-Time Markov Chains}

\mlsyssetsymbol{equal}{*}
\mlsyssetsymbol{amazon}{\ensuremath{\dagger}}

\begin{mlsysauthorlist}
\mlsysauthor{Siva Rajesh Kasa}{equal,amazon}
\mlsysauthor{Yasong Dai}{equal,anu}
\mlsysauthor{Sumit Negi}{amazon}
\mlsysauthor{Hongdong Li}{anu}
\end{mlsysauthorlist}

\mlsysaffiliation{anu}{School of Computing, Australian National University}

\mlsyscorrespondingauthor{Siva Rajesh Kasa}{kasa@u.nus.edu}
\mlsyscorrespondingauthor{Yasong Dai}{yasong.dai@anu.edu.au}

\mlsyskeywords{diffusion language models, continuous-time Markov chains, speculative sampling, exact simulation}

\vskip 0.3in

\begin{abstract}
    Masked diffusion language models offer parallel text generation, but independently sampling multiple tokens can disrupt their dependencies and change the output distribution. Exact continuous-time sampling avoids this approximation, yet typically requires one sequential model evaluation per token. We introduce SpecCTMC, a training-free speculative sampler that reduces this sequential bottleneck without a separate draft model. SpecCTMC proposes successive token updates using previously computed model predictions, evaluates their intermediate states in parallel, and applies an exact acceptance-and-correction procedure. We prove that SpecCTMC preserves the joint distribution of event times, generation order, and token values of the underlying continuous-time Markov chain. It therefore introduces neither time-discretization error nor additional factorization error from parallel decoding. Experiments on LLaDA-8B-Instruct and Dream-v0-Instruct-7B across five benchmarks demonstrate 1.73–4.38$\times$ fewer sequential model evaluations than direct continuous-time sampling, while preserving the reference sampling distribution.
    Our code is available at: \url{https://github.com/kasakh/specctmc}
\end{abstract}


]

\printAffiliationsAndNotice{\mlsysEqualContribution\quad\textsuperscript{\ensuremath{\dagger}}For Siva Rajesh Kasa and Sumit Negi, this work does not relate to their positions at Amazon.\quad}

\section{Introduction}
\label{sec:introduction}

Masked diffusion language models (MDLMs) generate text by iteratively unmasking tokens in a partially masked sequence, allowing generation beyond a fixed left-to-right order~\citep{sahoo2024simple,shi2024simplified,nie2025llada,ye2025dream}. This flexibility offers opportunities for parallel inference, but realizing them without compromising generation quality remains challenging. In the continuous-time formulation, the reverse process is a continuous-time Markov chain (CTMC) that unmasks one token at each event~\citep{campbell2022continuous}. Direct simulation therefore requires a sequence of model evaluations, with each update depending on previously unmasked tokens. Reducing this sequential cost is central to making exact sampling of diffusion language models practical.

\begin{figure}[t]
    \centering
    \includegraphics[width=1.1\linewidth]{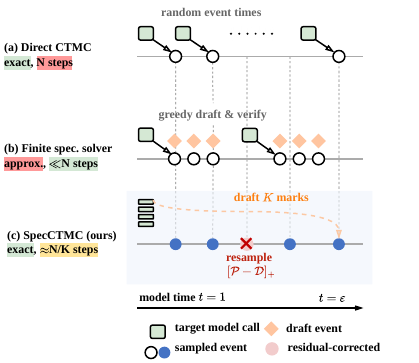}
    
    \caption{\textbf{Sampling a masked diffusion LLM as a CTMC.} (a) Direct event simulation is exact but requires one sequential target model evaluation per unmasking event. (b) Speculative dLLM decoders verify drafts against a finite-step greedy decoder, with \emph{restricted} lossless guarantees. (c) \method preserves the exact random event clock and the distribution of joint marks. The diagram's \(N/K\) label is the full-acceptance ideal when all drafted states fit in one target-model call; generally, one speculative round commits up to \(K\) events.}
    \label{fig:intro-overview}
\end{figure}

Existing approaches reduce inference cost through numerical approximation, decoding heuristics, and computation reuse. Numerical samplers such as $\tau$-leaping approximate multiple transitions within a time interval, while higher-order methods, including the $\theta$-trapezoidal solver, improve the accuracy of this approximation~\citep{campbell2022continuous,lou2024sedd,ren2025fast}. Practical decoders also use greedy token selection and confidence-based unmasking orders to generate high-quality text with fewer denoising steps~\citep{nie2025llada,ye2025dream}. Fast-dLLM combines confidence-guided parallel decoding with approximate KV caching~\citep{wu2025fastdllm}, and Fast-dLLM++ uses the full token-confidence profile to select larger groups of tokens for simultaneous commitment~\citep{kasa2026fastdllmpp}.

A central challenge in MDLMs is that tokens predicted from the same partially masked sequence are often not conditionally independent. Unmasking one token can change the distribution of another, so sampling several positions independently can result in a different distribution from updating them sequentially. This factorization error underlies the curse of parallel decoding discussed in previous works like~\citet{wu2025fastdllm,kasa2026fastdllmpp}.
Separately, the denoiser’s conditional predictions across different masking contexts need not be compatible with a single joint distribution \citep{young2024inconsistencies,torroba-hennigen-kim-2023-deriving}.

The choice of decoding policy also matters when generation is used to explore multiple candidate answers. Greedy decoding can be effective for producing a single answer, but a fully deterministic decoder cannot explore alternative solutions through repeated runs. Even with stochastic token sampling, confidence-based unmasking orders can restrict diversity and limit improvements from additional samples~\citep{olausson2026twotemperatures}. For reasoning and code generation, this motivates evaluating solution coverage across multiple attempts alongside single-sample accuracy, and accelerating stochastic generation while preserving its sampling distribution.

Exact event-driven simulation provides a principled reference for this goal. Gillespie's stochastic simulation algorithm samples individual transitions at random event times~\citep{gillespie1977exact}, and sequential first-hitting sampling applies this principle to absorbing-mask diffusion~\citep{zheng2025fhs}. After each position is unmasked, the model conditions the next update on the resulting sequence. This introduces neither time-discretization error nor the additional factorization error caused by committing independent token updates together. Its computational cost, however, remains dominated by sequential model evaluations.

To address these key bottlenecks, we introduce \textsc{SpecCTMC}, a training-free speculative sampler that reduces this sequential evaluations without the need for a separate draft model. Our approach builds on distribution-preserving speculative sampling for autoregressive (AR) language models~\citep{leviathan2023fast,chen2023speculative} and its extension to continuous diffusion models~\citep{debortoli2025accelerated}. Speculative methods have also been developed for finite-step diffusion language model decoders~\citep{agrawal2026spiffy,pan2026trajectory}. Here, we instead target the underlying absorbing-mask CTMC, including its random event times and the joint selection of token positions and values.

As shown in \Cref{fig:intro-overview}, the key observation is that, for the absorbing-mask generators considered here, event times depend on diffusion time and the number of remaining masks, independently of the identities of unmasked tokens. These times can therefore be sampled in advance and preserved when a proposed token update is rejected and corrected. \textsc{SpecCTMC} reuses previously computed model predictions to propose successive updates, evaluates their intermediate sequences in parallel, and applies an exact acceptance-and-correction procedure. We prove that the resulting trajectory has the same joint distribution of event times, token positions, and token values as direct simulation. 

Experiments on LLaDA-8B-Instruct and Dream-v0-Instruct-7B across five benchmarks show that \textsc{SpecCTMC} uses 1.73-4.38$\times$ fewer sequential target-model evaluations than direct simulation, with measured wall-clock speedups of up to 2.88$\times$ on an NVIDIA B200. On GSM8K, multiple stochastic attempts solve questions missed by deterministic greedy denoising, and \textsc{SpecCTMC} reduces their cumulative sequential-evaluation cost relative to DirectCTMC. As shown in \Cref{fig:intro-tradeoff}, by reducing the number of sequential evaluations, \textsc{SpecCTMC} addresses a cost complementary to optimizations that make each evaluation cheaper. In particular, cached or otherwise approximate predictions can be used for drafting while exact target verification preserves the CTMC sampling distribution.

Our contributions are:
\begin{itemize}
    \item We develop a training-free speculative sampler without a separate draft model
    for absorbing-mask diffusion that verifies multiple successive token
    updates through parallel model evaluation.

    \item We prove preservation of the complete CTMC trajectory
    distribution under the exactness conditions and, for an adaptive-depth
    variant under bounded unmasking influence, derive explicit bounds on
    sequential target batches and expected target work.

    \item We evaluate \textsc{SpecCTMC} across reasoning, coding, and
    instruction-following benchmarks, demonstrating reductions in
    sequential model evaluations, practical latency gains, and the
    benefits of stochastic generation for multiple-answer coverage.
\end{itemize}

\definecolor{specorange}{HTML}{FE8010}
\definecolor{taublue}{HTML}{468FD8}
\definecolor{trapgreen}{HTML}{70AD47}
\definecolor{directgray}{HTML}{404040}

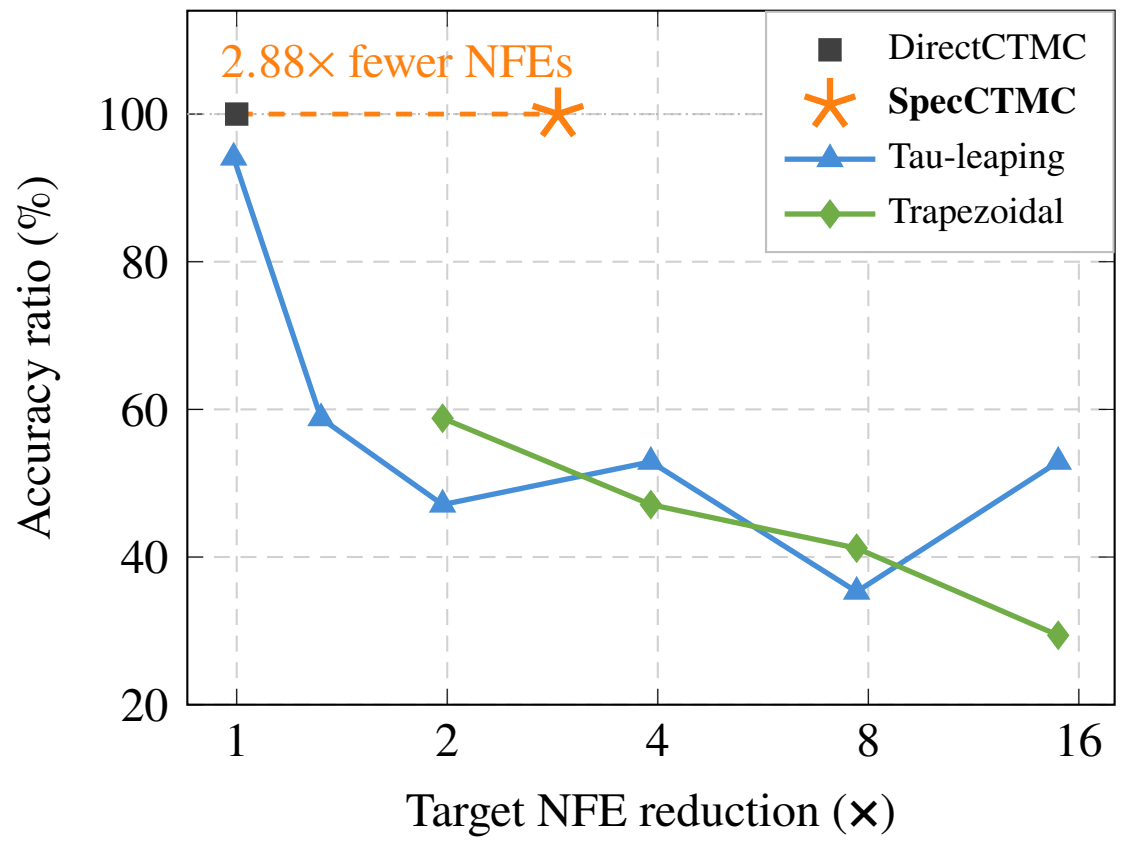
\begin{figure}[t]

\begin{minipage}[c]{0.45\textwidth}
\centering
\begin{tikzpicture}
\begin{axis}[
    width=\linewidth,
    height=0.8\linewidth,
    xlabel={Target NFE reduction ($\times$)},
    ylabel={Accuracy ratio (\%)},
    label style={font=\small\boldmath},
    tick label style={font=\footnotesize},
    xmode=log,
    log basis x=2,
    xmin=0.85, xmax=18,
    xtick={1,2,4,8,16},
    xticklabels={1,2,4,8,16},
    ymin=20, ymax=114,
    ytick={20,40,60,80,100},
    minor x tick num=0,
    axis line style={black},
    tick style={black},
    tick align=inside,
    major tick length=3pt,
    ymajorgrids=true,
    xmajorgrids=true,
    grid style={densely dashed,black!18},
    clip=false,
    legend style={
        at={(1.00,1.00)},
        anchor=north east,
        legend columns=1,
        draw=black!25,
        fill=white!50,
        font=\scriptsize,
        /tikz/every even column/.append style={column sep=3pt},
    },
    legend cell align=left,
]

\addplot[
    gray!55,
    densely dotted,
    thin,
    forget plot,
] coordinates {(0.85,100) (18,100)};

\addplot[
    only marks,
    color=directgray,
    mark=square*,
    mark size=2pt,
] coordinates {(1,100)};
\addlegendentry{DirectCTMC}

\addplot[
    only marks,
    color=specorange,
    mark=star,
    mark size=5pt,
    mark options={solid,line width=1.2pt},
] coordinates {(2.88,100)};
\addlegendentry{\bf SpecCTMC}

\draw[color=specorange,dashed,line width=0.9pt,]
    (axis cs:1,100) --
    node[midway,above=3pt,font=\footnotesize,text=specorange]
        {$2.88\times$ fewer NFEs}
    (axis cs:2.88,100);

\addplot[
    color=taublue,
    solid,
    line width=1pt,
    mark=triangle*,
    mark size=2pt,
    point meta=explicit symbolic,
] table[
    x=reduction,
    y=retained,
    meta=steps,
    row sep=\\,
] {
    reduction retained steps \\
    0.99      94.1     256   \\
    1.32      58.8     192   \\
    1.97      47.1     128   \\
    3.91      52.9     64    \\
    7.70      35.3     32    \\
    14.95     52.9     16    \\
};
\addlegendentry{Tau-leaping}

\addplot[
    color=trapgreen,
    solid,
    line width=1pt,
    mark=diamond*,
    mark size=2pt,
    point meta=explicit symbolic,
] table[
    x=reduction,
    y=retained,
    meta=steps,
    row sep=\\,
] {
    reduction retained steps \\
    1.97      58.8     64    \\
    3.91      47.1     32    \\
    7.70      41.2     16    \\
    14.95     29.4     8     \\
};
\addlegendentry{Trapezoidal}

\end{axis}
\end{tikzpicture}
\end{minipage}

\caption{
{\bf Exact law preserving of \method on NFE acceleration.} On GSM8K with LLaDA-8B-Instruct (256 generated tokens), SpecCTMC reduces target NFEs by $2.88\times$ relative to DirectCTMC while preserving expected accuracy. Finite-step samplers trade accuracy for fewer evaluations.
}

\label{fig:intro-tradeoff}
\end{figure}
\section{Related Work}
\label{sec:related}

\paragraph{Discrete diffusion simulation.}
Discrete diffusion admits a continuous-time Markov chain formulation with random state transitions~\citep{campbell2022continuous}. For absorbing-mask models, sequential first-hitting sampling draws each unmasking time and then updates one token~\citep{zheng2025fhs}; DirectCTMC instantiates this exact event-driven reference. Parallel first-hitting sampling uses the sampled event times but freezes token predictions across several updates, so exact clocks alone do not preserve the transition law. Uniformization instead obtains exact simulation from a common candidate clock and virtual self-transitions~\citep{chen2024uniformization}, whereas tau-leaping and higher-order solvers approximate the dynamics over finite intervals~\citep{ren2025fast}. SpecCTMC evaluates tentative future states together while preserving the joint law of event times, positions, and tokens.

\paragraph{Parallel and speculative masked decoding.}
Parallel masked decoders reduce model evaluations by committing several tokens from one set of predictions. Fast-dLLM and Fast-dLLM++ select updates by confidence, EB-Sampler uses entropy bounds, and PUNT tests approximate conditional independence~\citep{wu2025fastdllm,kasa2026fastdllmpp,benhamu2025ebsampler,azangulov2025punt}. Schedule analyses characterize the resulting dependency error, while DEMASK and SIRS use learned influence estimates or rejection sampling to control it~\citep{chen2025schedules,ringel2026demask,zhang2026sirs}. Self Speculative Decoding reproduces a deterministic stepwise reference; Spiffy, Parallel Speculative Decoding, and trajectory-level speculation draft future denoising states under method-specific exactness conditions or measured speed and quality tradeoffs~\citep{gao2025ssd,agrawal2026spiffy,sun2026psd,pan2026trajectory}. Self-Speculative Masked Diffusions instead train a hybrid causal and noncausal architecture~\citep{campbell2026self}. SpecCTMC uses previously computed predictions only to draft tentative marks. Target verification and positive-residual correction then preserve DirectCTMC without additional training or a separate drafter.

\paragraph{Exact speculative and parallel sampling.}
Exact speculative sampling has been developed for continuous diffusion transitions on fixed denoising schedules, with block verification improving progress across proposed transitions~\citep{debortoli2025accelerated,soen2026block}. AutoSpec uses exchangeable increments to parallelize continuous diffusion without a separate draft model~\citep{hu2025autospec}. Autospeculation and Any-Subset Speculative Decoding obtain exact parallel samples from compatible conditional-marginal or joint-density oracles, and related theory parallelizes single-site dynamics under influence conditions~\citep{anari2025autospeculation,guo2026anysubset,liu2024parallelize}. Temporal-point-process methods likewise draft and verify marked event sequences~\citep{gong2025tppsd,bilos2025tpp}. SpecCTMC requires normalized state-dependent transition rows, but not globally compatible denoiser conditionals. For absorbing-mask models, correcting a proposed position and token leaves the future event clock valid, which permits exact speculation over the complete marked CTMC path. \Cref{tab:method-positioning} summarizes these distinctions; \cref{app:related-work} gives the detailed comparisons.

\begin{table*}[t]
\centering
\caption{Comparison of masked-diffusion inference methods. ``Natural-event CTMC law'' is the stochastic process induced by the unchanged denoiser. ``Within-trajectory parallelism'' means that one target round can advance multiple events or positions.}
\label{tab:method-positioning}
\scriptsize
\setlength{\tabcolsep}{1.8pt}
\renewcommand{\arraystretch}{1.09}
\resizebox{\textwidth}{!}{%
\begin{tabular}{@{}lcccccc@{}}
\toprule
Method & \shortstack{Training\\free} & \shortstack{Exact event\\times} & \shortstack{Natural-event\\CTMC law} & \shortstack{No separate\\drafter} & \shortstack{Within-trajectory\\parallelism} & \shortstack{No added\\factorization error} \\
\midrule
\multicolumn{7}{@{}l}{\textit{Continuous-time Markov chain simulation}} \\
Fixed-step ancestral / tau-leaping~\citep{campbell2022continuous,ren2025fast} & Yes & No & No & Yes & Yes & No \\
Sequential FHS (DirectCTMC)~\citep{zheng2025fhs} & Yes & Yes & Yes & Yes & No & Yes \\
Parallel FHS~\citep{zheng2025fhs} & Yes & Yes & No & Yes & Yes & No \\
\midrule
\multicolumn{7}{@{}l}{\textit{Parallel masked-diffusion decoding}} \\
Fast-dLLM~\citep{wu2025fastdllm} & Yes & No & No & Yes & Yes & No \\
Fast-dLLM++~\citep{kasa2026fastdllmpp} & Yes & No & No & Yes & Yes & No \\
EB-Sampler~\citep{benhamu2025ebsampler} & Yes & No & No & Yes & Yes & No \\
PUNT~\citep{azangulov2025punt} & Yes & No & No & Yes & Yes & No \\
\midrule
\multicolumn{7}{@{}l}{\textit{Speculative masked-diffusion decoding}} \\
Spiffy, practical LLaDA/Dream variant~\citep{agrawal2026spiffy} & Yes & No & No & Yes & Yes & No \\
Self Speculative Decoding~\citep{gao2025ssd} & Yes & No & No & Yes & Yes & Yes \\
Parallel Speculative Decoding~\citep{sun2026psd} & Yes & No & No & Yes & Yes & No \\
Trajectory-Level Speculative Decoding, practical variant~\citep{pan2026trajectory} & Yes & No & No & Yes & Yes & No \\
Self-Speculative Masked Diffusions~\citep{campbell2026self} & No & No & No & Yes & Yes & Yes \\
\midrule
\multicolumn{7}{@{}l}{\textit{Exact speculative natural-event simulation}} \\
\textbf{SpecCTMC (ours)} & \textbf{Yes} & \textbf{Yes} & \textbf{Yes} & \textbf{Yes} & \textbf{Yes} & \textbf{Yes} \\
\bottomrule
\end{tabular}%
}
\end{table*}

\section{Event-Driven Masked Diffusion}
\label{sec:problem}

Given an absorbing-mask diffusion model like~\cite{ye2025dream,nie2025llada}, we define a target continuous-time Markov chain (CTMC) from its predicted token probabilities at masked positions and the masking schedule. We call its exact event-driven sampler DirectCTMC. Unlike decoding on a finite-step time grid, DirectCTMC samples the next event time from the CTMC's total exit intensity. Then it samples a position-token mark and unmasks exactly one position.

\paragraph{State and marks.}
Let \(\cX\) be the finite set of token sequences for one generation block, and let $\mask$ be the mask token. For state \(x\in\cX\), define the masked positions:
\begin{equation}
A(x):=\{d:x_d=\mask\}, \quad m(x)=|A(x)|.
\end{equation}
A marked event is then defined as \(z=(d,v)\), where \(d\in A(x)\) is a masked position and \(v\neq\mask\) is its replacement token. For a valid mark \(z=(d,v)\), define \(\Phi(x,z)\) to be the sequence equal to \(x\) at every coordinate except \(d\), with
\begin{equation}
  [\Phi(x,(d,v))]_d=v.
\end{equation}

\paragraph{Target generator.}
Let \(p_\theta(v\mid x,d)\) be the normalized target probability for replacement token \(v\) at masked position \(d\) for a fixed model \(\theta\). The model defines its unmasking process running backward in model time from \(t=1\) to a positive cutoff \(\varepsilon\). The LLaDA, Dream, and LLaDA-V formulations used here mask each generated-answer token independently with probability \(t\) at model time \(t\)~\citep{nie2025llada,ye2025dream,you2025lladav}. \Cref{app:masking-schedule} derives the resulting \(1/t\) factor and gives the corresponding expression for a different monotone masking schedule. For these models, the event rate of the underlying CTMC is
\begin{equation}
q_t(x,(d,v))=\frac{1}{t}p_\theta(v\mid x,d).
\label{eq:generator}
\end{equation}
For each masked position \(d\), \(p_\theta(\cdot\mid x,d)\) is normalized over replacement tokens \(v\neq\mask\), so the total exit intensity is
\begin{equation}
\lambda_t(x)=\sum_{d\in A(x)}\sum_{v\neq\mask}q_t(x,(d,v))
=\frac{m(x)}{t}.
\label{eq:total-rate}
\end{equation}
The target mark distribution conditioned on an event is:
\begin{equation}
P_{t,x}(d,v)=\frac{1}{m(x)}p_\theta(v\mid x,d).
\label{eq:mark-kernel}
\end{equation}
Thus, the position is uniformly distributed over masked positions, while the token follows the model probability at that position.

\paragraph{Exact event clock.}
Suppose the current time is \(t\), the current state is \(x\), and \(m(x)>0\). For \(E\sim\operatorname{Exponential}(1)\), define
\begin{equation}
T_{\mathrm{next}}=t\cdot\exp\left(-\frac{E}{m(x)}\right).
\label{eq:event-clock}
\end{equation}
This transformation has the exact survival law associated with \(\lambda_s(x)=m(x)/s\). Every event unmasks exactly one mask, so the future intensity depends on the number of events, but not on the identities of their marks. The process has at most the initial number of masks and is therefore non-explosive. If the next time falls below the cutoff \(\varepsilon\), simulation stops and both methods deterministically fill the remaining masked positions using the same rule.
\paragraph{Parallel commitment and conditional dependence.}
Unmasking several tokens from the same state can change their joint distribution because the sampled tokens do not condition on one another.

We illustrate this side-effect with an example of $n=2r$ binary positions arranged in $r$ independent pairs. Each pair follows the distribution:
\begin{equation}
\begin{aligned}
  Q_a(00)=Q_a(11)&=\frac{a}{2},\\
  Q_a(01)=Q_a(10)&=\frac{1-a}{2},\qquad a\in(0,1].
\end{aligned}
  \label{eq:pair-target}
\end{equation}
For \(x=(x_1,\ldots,x_{2r})\), the full target distribution over the \(r\) independent pairs is \(Q_a^{\otimes r}(x)=\prod_{j=1}^{r}Q_a(x_{2j-1},x_{2j})\). Assume the model returns its exact conditional token distributions given the unmasked tokens.

Consider a general parallel decoder that independently assigns each position to a denoising step $s\in\{1,\ldots,T\}$ with probability $w_s$. At each step with at least one assigned position, the decoder evaluates the current state once and independently samples all assigned positions from their conditional token distributions. Here we assume both DirectCTMC and parallel decoder run until all positions are unmasked, i.e., without stopping at the cutoff $\varepsilon$. Under these assumptions, we obtain \Cref{lem:parallel-dependence} as follows:

\begin{lemma}[Parallel commitment attenuates dependence]
\label{lem:parallel-dependence}
Let $S:=\sum_{s=1}^T w_s^2$ be the probability that two token positions are assigned to the same iteration. DirectCTMC returns $Q_a^{\otimes r}$, whereas the parallel decoder returns $Q_{a_T}^{\otimes r}$, where
\begin{equation}
  a_T=(1-S)a+\frac{S}{2}
      =a-\left(a-\frac12\right)S.
  \label{eq:parallel-attenuation}
\end{equation}
Consequently, the probabilities that all $r$ pairs agree are $a^r$ and $a_T^r$, respectively.

When $a=1$, requiring all pairs to agree with probability at least $1-\delta,\; \delta\in(0,1)$ implies
\begin{equation*}
\begin{aligned}
  T&\ge \frac{n}{4\log(1/(1-\delta))},\mathbb E[M]\ge n+2(n-1)\log{1-\delta},
\end{aligned}
\label{eq:parallel-evaluation-bound}
\end{equation*}
where $M$ is the number of denoising steps with at least one assigned position, hence the number of target-model evaluations.
\end{lemma}

\Cref{eq:parallel-attenuation} shows that independent parallel sampling undermines the pairwise dependence compared with target distribution. For a setting $a=1, \;\delta=0.05$, the parallel decoder requires at least approximately $0.897n$ target-model evaluations on average.

While the above example identifies the loss of pairwise dependence caused by independent parallel sampling, \method batches target-model evaluations at successive tentative states and verifies the proposed marks sequentially. The verification with residual correction preserves the DirectCTMC distribution while allowing several target evaluations to share one batched evaluation.

\textbf{Exact reference simulation.}
DirectCTMC repeatedly samples \(T_{\mathrm{next}}\) with \cref{eq:event-clock}, evaluates \(P_{T_{\mathrm{next}},x}\), samples one position-token mark, and commits it. It therefore uses one sequential target-model round per natural event and defines the reference path law. By contrast, tau-leaping and trapezoidal solvers \emph{approximate the CTMC} by freezing or extrapolating transition rates over finite time intervals~\citep{campbell2022continuous,ren2025fast}. Our goal is to reduce sequential target-model evaluations while preserving the reference path law.

\section{Exact Speculative CTMC Simulation}
\label{sec:method}


\definecolor{specblue}{HTML}{1D4ED8}
\definecolor{verifred}{HTML}{C0162F}
\providecommand{\spec}[1]{{\color{specblue}#1}}
\providecommand{\verif}[1]{{\color{verifred}#1}}

\begin{algorithm}[t]
\caption{\\ Exact speculative simulation of one marked CTMC block.
}
\label{alg:speculative-ctmc}
\begin{algorithmic}[1]
\REQUIRE Initial state $X_0$, time $t$, cutoff $\varepsilon$, \spec{depth $K$}
\STATE $m_0 \gets |A(X_0)|$, \quad $T_0 \gets t$, \quad $n \gets 0$
\STATE Draw $E_j \overset{\mathrm{iid}}{\sim}\operatorname{Exp}(1)$, $j=1,\ldots,m_0$
\STATE $T_j \gets T_{j-1}\cdot\exp\!\bigl(-E_j/(m_0-j+1)\bigr)$, $j=1,\ldots,m_0$
\STATE $N \gets |\{j\in\{1,\ldots,m_0\}:T_j\ge\varepsilon\}|$
\WHILE{$n<N$}
    \STATE \spec{$L \gets \min\{K,N-n\}$, \quad $\widehat X_0 \gets X_n$}
    \FOR{\spec{$i=1,\ldots,L$}}
        \STATE \spec{Form $D_i$ on $\mathcal Z(\widehat X_{i-1})$; draw $\widehat Z_i\sim D_i$}
        \STATE \spec{$\widehat X_i \gets \Phi(\widehat X_{i-1},\widehat Z_i)$}
    \ENDFOR
    \STATE \spec{$P_i \gets P_{T_{n+i},\widehat X_{i-1}}$, $i=1,\ldots,L$}
           \hfill \spec{\footnotesize(batched)}
    \FOR{$i=1,\ldots,L$}
        \STATE \verif{Draw $U_i\sim\operatorname{Unif}(0,1)$}
        \IF{\verif{$U_i\le a_i(\widehat Z_i)$}}
            \STATE \verif{$Z_{n+i}\gets\widehat Z_i$}
        \ELSE
            \STATE \verif{Draw $Z_{n+i}\sim R_i$} \hfill \verif{\footnotesize(\cref{eq:residual})}
        \ENDIF
        \STATE $X_{n+i}\gets \Phi(\widehat X_{i-1},Z_{n+i})$, \quad \verif{$\ell\gets i$}
        \IF{\verif{$U_i>a_i(\widehat Z_i)$}}
            \STATE \verif{\textbf{break}}
        \ENDIF
    \ENDFOR
    \STATE $n\gets n+\verif{\ell}$
\ENDWHILE
\STATE Apply the common deterministic completion rule to $X_N$
\end{algorithmic}
\end{algorithm}

\method applies exact speculative decoding to the event-driven masked diffusion process described in \Cref{sec:problem}. As previously established, the event time distribution depends only on model time and the number of remaining masks. \method therefore samples all event times before generating their marks and speculates only over the position-token marks.
Each speculative round drafts a tentative trajectory, evaluates the target distributions at the states before the proposed updates, and verifies the proposed marks in sequence. In our implementation, the drafter selects a masked position uniformly and samples its replacement token from a mixture of the previously computed target probabilities at that position and a uniform distribution over tokens other than the mask token. \Cref{app:reproducibility} gives the mixture weight and cache details. At the first rejection, \method samples a corrected mark from the residual correction distribution and starts a new round from the corrected state.



\paragraph{Batched target evaluation.}
The tentative trajectory provides the states before the proposed updates $\widehat X_0,\ldots,\widehat X_{L-1}$ needed to evaluate
\begin{equation}
P_i=P_{T_{n+i},\widehat X_{i-1}}, \quad i=1,\ldots,L,
\end{equation}
We evaluate these distributions in one or more target-model calls, reusing any target distribution already available for the same state. \Cref{app:protocol} states how target states are grouped for each experiment.

Each input row represents the state before one proposed update. The target distribution at index \(i\) is conditioned on the preceding proposed updates. A verification that reaches index $i$ means those proposals have all been accepted, so $\widehat X_{i-1}$ equals the current committed state. The target row therefore uses the same conditioning state as DirectCTMC.

\paragraph{Sequential verification and correction.}
For a proposed mark $\widehat Z_i=z$, define
\begin{equation}
    a_i(z)
    =
    \min\left\{1,\frac{P_i(z)}{D_i(z)}\right\}.
    \label{eq:acceptance}
\end{equation}
A sampled proposal has $D_i(z)>0$. Draw $U_i\sim\operatorname{U}(0,1)$ independently and accept the proposal when $U_i\le a_i(\widehat Z_i)$.
Averaging over the proposed mark gives $\alpha_i = \sum_z \min\{P_i(z),D_i(z)\}$. At the first rejection, sample a corrected mark from the residual correction distribution
\begin{equation}
    R_i(z) = \frac{\pospart{P_i(z)-D_i(z)}}{1-\alpha_i}.
    \label{eq:residual}
\end{equation}
where the denominator is the rejection probability and is positive whenever rejection occurs. If $\alpha_i=1$, acceptance is certain and no residual distribution will be formed.

The corrected mark is committed at time $T_{n+i}$. SpecCTMC then discards the remaining tentative marks, states, and target distributions, and restarts drafting from the corrected state. If all proposals are accepted, the entire draft is committed.

\Cref{alg:speculative-ctmc} summarizes the procedure of \method, where \spec{Blue} denotes drafting and batched target evaluation; \verif{red} denotes sequential verification and residual correction. Black lines denote operations shared with DirectCTMC. \method preserves sequential conditioning between committed events while batching the target evaluations along each tentative trajectory.

\section{Theoretical Analysis}
\label{sec:theory}

We establish that \method preserves the joint distribution of event times, position-token marks, and resulting states generated by DirectCTMC. The argument has two parts: (1) correcting a proposed mark leaves the event-time distribution unchanged, and (2) verification with residual correction produces marks with the target conditional distribution. We then characterize the number of speculative rounds and target evaluations, and relate this cost to proposal quality.

\subsection{Event-Time Sampling}
\label{sec:event-time}

\paragraph{Validity of event times sampled before their marks.}
Let $m_0=m(X_0)$ and $T_0=1$ be the initial mask count and event time. After $j-1$ events, there will be exactly $m_0-j+1$ masked positions remaining. The total exit intensity is $\lambda_t(x)=(m_0-j+1)/t$, regardless of the previously unmasked positions and their token values.

We can thus sample the event times before generating any marks by:
\begin{equation}
  T_j=T_{j-1}\cdot\exp\!\left(-\frac{E_j}{m_0-j+1}\right),
  \quad j=1,\ldots,m_0.
  \label{eq:retained-clock}
\end{equation}
where the \(E_j\) are independent unit exponential random variables. Use the times at or above the cutoff \(\varepsilon\), and define
\begin{equation}
\begin{aligned}
  N&=\left|\{j\in\{1,\dots,m_0\}:T_j\geq\varepsilon\}\right|,\\
  \mathcal C&=(N,T_1,\ldots,T_N).
\end{aligned}
\end{equation}
We obtain the following lemma.

\begin{lemma}[Validity of sampled event times]
\label{lem:retained-clock}
Conditional on $X_0$, the clock $\mathcal C$ has the same distribution as the event count and event times generated by DirectCTMC. Replacing a proposed position-token mark by another valid mark preserves the validity of all remaining sampled event times.
\begin{equation}
\begin{aligned}
&\operatorname{Law}_{Spec}
  \!\left(N,(T_j)_{j=1}^{N}| X_0\right) =
\operatorname{Law}_{{Direct}}
  \!\left(N,(T_j)_{j=1}^{N}| X_0\right).
\end{aligned}
\end{equation}
\end{lemma}

It remains to show that verification and correction produce the target mark distribution at each sampled event time, conditioned on the current committed state.

\subsection{Exactness of the complete marked path}
\label{sec:path-exactness}

The conditions for equality of the joint event distributions appear in \Cref{app:exactness-assumptions}.

\begin{theorem}[Exact preservation of the joint event distribution]
\label{thm:marked-path}
Under Conditions C1 to C8 in \cref{app:exactness-assumptions}, \method and DirectCTMC satisfy
\begin{equation}
\begin{aligned}
 &\operatorname{Law}_{{Spec}}
   \left(X_0,N,(T_j,Z_j,X_j)_{j=1}^{N}\right)\\
 &\qquad=\operatorname{Law}_{{Direct}}
   \left(X_0,N,(T_j,Z_j,X_j)_{j=1}^{N}\right).
\end{aligned}
\label{eq:path-law-equality}
\end{equation}
After stopping at the cutoff $\varepsilon$, applying the same deterministic completion rule also gives equal output distributions.
\end{theorem}

Complete assumptions and proofs are given in \cref{app:proofs,app:additional-theory}.


By \Cref{thm:marked-path}, independent repetitions of \method and DirectCTMC have the same joint output distribution, and therefore the same pass@$k$ for every $k$. We will discuss this implication in \Cref{sec:experiments}.


\providecommand{\pty}[1]{\textcolor{black!45}{#1}} 
\providecommand{\slw}[1]{\textcolor{red!65!black}{#1}} 
\providecommand{\dpt}[1]{\,{\scriptsize\textcolor{black!55}{(#1)}}} 

\begin{table*}[t]
\centering
\caption{\textbf{Exact speculative CTMC sampling on LLaDA.} For each generated length $L$ we report wall-clock speedup of SpecCTMC over DirectCTMC (Wall-clock) and the reduction in sequential target-model calls (NFE); the draft depth used is in parentheses. Gray speedups are at parity (the paired 95\% interval contains one); all others are faster, and no condition is slower. The last row reports accuracy (\%) of DirectCTMC $\rightarrow$ SpecCTMC averaged over lengths; every paired 95\% confidence interval for the SpecCTMC-minus-DirectCTMC accuracy difference contains zero. Bold marks the best value.}
\label{tab:llada-compact}
\small
\setlength{\tabcolsep}{3pt}
\begin{tabular}{lcccccccccccc}
\toprule
 & \multicolumn{2}{c}{GSM8K} & \multicolumn{2}{c}{MATH} & \multicolumn{2}{c}{MBPP} & \multicolumn{2}{c}{HumanEval} & \multicolumn{2}{c}{IFEval} & \multicolumn{2}{c}{Avg.} \\
\cmidrule(lr){2-3}\cmidrule(lr){4-5}\cmidrule(lr){6-7}\cmidrule(lr){8-9}\cmidrule(lr){10-11}\cmidrule(lr){12-13}
$L$ & Wall-clock & NFE & Wall-clock & NFE & Wall-clock & NFE & Wall-clock & NFE & Wall-clock & NFE & Wall-clock & NFE \\
\midrule
32 & \pty{1.00$\times$} & 1.75\dpt{2} & 1.15$\times$ & 1.75\dpt{2} & \pty{1.10$\times$} & 2.39\dpt{3} & 2.19$\times$ & 3.02\dpt{5} & 2.65$\times$ & 3.42\dpt{7} & 1.62$\times$ & 2.46 \\
64 & \pty{1.01$\times$} & 1.83\dpt{2} & 1.18$\times$ & 1.85\dpt{2} & 1.05$\times$ & 1.85\dpt{2} & 2.48$\times$ & 3.88\dpt{6} & 2.77$\times$ & 3.56\dpt{5} & 1.70$\times$ & 2.59 \\
128 & 1.06$\times$ & 2.70\dpt{3} & 1.18$\times$ & 2.71\dpt{3} & 1.12$\times$ & 1.90\dpt{2} & 2.37$\times$ & 3.94\dpt{5} & \textbf{2.88}$\times$ & 4.02\dpt{5} & 1.72$\times$ & 3.05 \\
256 & 1.08$\times$ & 2.81\dpt{3} & 1.10$\times$ & 2.81\dpt{3} & 1.10$\times$ & 1.93\dpt{2} & 1.92$\times$ & 2.80\dpt{3} & 2.22$\times$ & \textbf{4.38}\dpt{5} & 1.48$\times$ & 2.95 \\
512 & 1.09$\times$ & 3.81\dpt{4} & 1.07$\times$ & 2.89\dpt{3} & 1.07$\times$ & 1.96\dpt{2} & 1.38$\times$ & 2.87\dpt{3} & 1.62$\times$ & 2.91\dpt{3} & 1.25$\times$ & 2.89 \\
\midrule
Avg. & 1.05$\times$ & 2.58 & 1.14$\times$ & 2.40 & 1.09$\times$ & 2.01 & 2.07$\times$ & 3.30 & 2.43$\times$ & 3.66 & 1.55$\times$ & 2.79 \\
Acc. & \multicolumn{2}{c}{27.1 $\rightarrow$ 28.3} & \multicolumn{2}{c}{13.2 $\rightarrow$ 12.6} & \multicolumn{2}{c}{5.4 $\rightarrow$ 6.5} & \multicolumn{2}{c}{5.2 $\rightarrow$ 6.0} & \multicolumn{2}{c}{29.0 $\rightarrow$ 28.8} & \multicolumn{2}{c}{16.0 $\rightarrow$ 16.5} \\
\bottomrule
\end{tabular}
\end{table*}
\begin{table*}[t]
\centering
\caption{
{\bf Exact speculative CTMC sampling on Dream.} For each generated length $L$ we report the wall-clock speedup of SpecCTMC over DirectCTMC (Wall-clock) and the reduction in sequential target model function evaluations (NFE); depth used shown in \dpt{}. The last row reports accuracy (\%) of DirectCTMC $\rightarrow$ SpecCTMC, averaged over lengths.
}
\label{tab:dream-compact}
\footnotesize
\setlength{\tabcolsep}{2pt}
\begin{tabular}{lcccccccccccc}
\toprule
 & \multicolumn{2}{c}{GSM8K} & \multicolumn{2}{c}{MATH} & \multicolumn{2}{c}{MBPP} & \multicolumn{2}{c}{HumanEval} & \multicolumn{2}{c}{IFEval} & \multicolumn{2}{c}{Avg.} \\
\cmidrule(lr){2-3}\cmidrule(lr){4-5}\cmidrule(lr){6-7}\cmidrule(lr){8-9}\cmidrule(lr){10-11}\cmidrule(lr){12-13}
$L$ & Wall-clock & NFE & Wall-clock & NFE & Wall-clock & NFE & Wall-clock & NFE & Wall-clock & NFE & Wall-clock & NFE \\
\midrule
32 & \pty{0.91$\times$} & 1.73$\times$\dpt{2} & \pty{0.99$\times$} & 1.74$\times$\dpt{2} & \pty{0.96$\times$} & 1.75$\times$\dpt{2} & 1.94$\times$ & 2.67$\times$\dpt{4} & 1.80$\times$ & 2.87$\times$\dpt{5} & 1.32$\times$ & 2.15$\times$ \\
64 & \pty{0.96$\times$} & 1.83$\times$\dpt{2} & \pty{1.02$\times$} & 1.84$\times$\dpt{2} & \pty{1.03$\times$} & 1.83$\times$\dpt{2} & 1.97$\times$ & 3.46$\times$\dpt{5} & 2.62$\times$ & 3.72$\times$\dpt{6} & 1.52$\times$ & 2.54$\times$ \\
128 & \pty{0.99$\times$} & 1.89$\times$\dpt{2} & 1.05$\times$ & 2.71$\times$\dpt{3} & 1.02$\times$ & 2.74$\times$\dpt{3} & 2.00$\times$ & 2.69$\times$\dpt{3} & 2.46$\times$ & 3.97$\times$\dpt{5} & 1.50$\times$ & 2.80$\times$ \\
256 & \pty{0.99$\times$} & 1.95$\times$\dpt{2} & \pty{1.00$\times$} & 2.82$\times$\dpt{3} & 1.03$\times$ & 3.72$\times$\dpt{4} & 1.61$\times$ & 4.32$\times$\dpt{5} & 1.96$\times$ & 3.63$\times$\dpt{4} & 1.32$\times$ & 3.29$\times$ \\
512 & \pty{1.00$\times$} & 2.94$\times$\dpt{3} & \pty{1.00$\times$} & 3.82$\times$\dpt{4} & \pty{1.02$\times$} & 3.88$\times$\dpt{4} & 1.23$\times$ & 3.82$\times$\dpt{4} & 1.33$\times$ & 2.91$\times$\dpt{3} & 1.12$\times$ & 3.48$\times$ \\
\midrule
Avg. & 0.97$\times$ & 2.07$\times$ & 1.01$\times$ & 2.59$\times$ & 1.01$\times$ & 2.79$\times$ & 1.75$\times$ & 3.39$\times$ & 2.03$\times$ & 3.42$\times$ & 1.36$\times$ & 2.85$\times$ \\
Acc. & \multicolumn{2}{c}{20.5 $\rightarrow$ 20.6} & \multicolumn{2}{c}{6.6 $\rightarrow$ 6.9} & \multicolumn{2}{c}{8.8 $\rightarrow$ 9.3} & \multicolumn{2}{c}{7.8 $\rightarrow$ 7.4} & \multicolumn{2}{c}{30.3 $\rightarrow$ 28.6} & \multicolumn{2}{c}{14.8 $\rightarrow$ 14.6} \\
\bottomrule
\end{tabular}
\end{table*}

\paragraph{Target-model evaluations and latency.}
Let $b$ be the number of input rows per target model evaluation, and $I_r\in\{0,1\}$ an indicator function on whether round $r$ reuses its first target distribution. Evaluating the full draft requires $Q_r=\left\lceil\frac{L_r-I_r}{b}\right\rceil$ target-model calls, excluding initialization and deterministic completion at the cutoff.
The total number of newly evaluated states is
\begin{equation}
  W_{\mathrm{eval}}
  =B+F+\sum_{r=1}^{R}(L_r-I_r),
  \label{eq:evaluated-rows}
\end{equation}
where $B$ and $F$ count states evaluated for initialization and deterministic completion at the cutoff, respectively. These counts distinguish proposed events, evaluated states, and target-model calls. Latency additionally depends on batch evaluation cost, drafting, and verification, as we will discuss in \Cref{sec:experiments}.

\section{Experiments}
\label{sec:experiments}

\paragraph{Model and benchmark.}
We evaluate LLaDA-8B-Instruct and Dream-v0-Instruct-7B on GSM8K~\citep{cobbe2021gsm8k}, MATH~\citep{hendrycks2021math}, MBPP~\citep{austin2021mbpp}, HumanEval~\citep{chen2021humaneval}, and IFEval~\citep{zhou2023ifeval}, reporting sequential target-model evaluations, accuracy, and wall-clock time. We also evaluate the multimodal diffusion language model LLaDA-V~\citep{you2025lladav} on MathVerse~\citep{zhang2024mathverse} and MathVista~\citep{lu2024mathvista}, reporting NFE reduction and acceptance.

\paragraph{Baselines.}
We compare \method with DirectCTMC, the exact event-driven sampler, and two approximate samplers, Tau-leaping and trapezoidal sampling~\citep{campbell2022continuous,lou2024sedd}. These comparisons assess the reduction in target-model evaluations under exact sampling and the cost-quality trade-off of finite-step approximate samplers.

\paragraph{Implementation details.}
We evaluate generation lengths \(L\in\{32,64,128,256,512\}\) at sampling temperature one. The speculation depth is selected separately for each condition from \(\{2,\ldots,8\}\) by warmed timing on five fixed questions. Acceptance rate is the fraction of drafted events accepted during verification. We report wall-clock speedup and the reduction in sequential target-model evaluations (NFE), both relative to DirectCTMC. Unless stated otherwise, wall-clock time is measured on one NVIDIA B200 GPU. Appendix~\ref{app:experiments} gives the depth-selection and evaluation protocols.

\subsection{Reduced Target Evaluation against DirectCTMC}

Results are averaged across the fixed evaluation set and sampling seeds. The NFE reduction is defined as $R_{\text{NFE}}=N_{Direct}/N_{Spec}$, where \(N_{\mathrm{Direct}}\) and \(N_{\mathrm{Spec}}\) are the mean sequential target-model evaluations for DirectCTMC and SpecCTMC.

\begin{figure*}[t]
\centering
\begingroup
\definecolor{fwdb200}{HTML}{1F77B4}
\definecolor{fwdl40s}{HTML}{FF7F0E}
\definecolor{dephumaneval}{HTML}{D55E00}
\definecolor{depgsm}{HTML}{0072B2}

\pgfplotstableread{
n     lladaB200 lladaL40S dreamB200 dreamL40S
400   1.3347    2.5853    1.3762    2.0170
450   1.4576    2.4369    1.5517    2.2818
500   1.5745    2.8022    1.6834    2.4517
550   1.7010    2.5653    1.8389    2.4943
600   1.9078    2.7818    1.9857    2.7330
650   2.0262    2.4690    2.1058    2.3950
700   2.1369    2.5484    2.2085    2.5080
800   2.4887    2.9267    2.5361    2.6400
900   2.6216    3.1420    2.6124    2.6431
1000  2.6693    3.2605    2.6664    2.8158
1100  2.6391    2.7025    2.7434    2.9341
1200  2.7225    3.0119    2.8515    2.9465
1350  2.7501    3.0717    2.7509    2.8926
1500  2.7898    3.2367    2.8159    2.9875
2000  2.9071    3.3556    3.0779    3.1119
2500  2.9292    3.2693    3.1677    3.2216
3100  2.8840    3.2978    3.0070    3.1901
}\fwdcosttable

\pgfplotstableread{
d  heB200 heB200lo heB200hi heL40S heL40Slo heL40Shi gsmB200 gsmB200lo gsmB200hi
2  1.3730 1.3615   1.4172   1.1727 1.0600   1.1905   0.9925  0.9830    1.0111
3  1.4936 1.4348   1.5631   1.1804 1.1312   1.2223   0.9969  0.9522    1.0148
4  1.4303 1.3956   1.5955   1.1200 1.0006   1.2149   0.9411  0.9096    1.0140
5  1.6135 1.4622   1.7696   1.0760 1.0664   1.2360   0.9121  0.8341    0.9897
6  1.4806 1.3981   1.6007   1.0536 0.9575   1.1524   0.8928  0.8727    0.9750
7  1.3626 1.2747   1.5085   0.9829 0.9248   1.0442   0.8514  0.7975    0.9616
8  1.3639 1.1086   1.4156   0.9425 0.8751   1.0454   0.7940  0.7466    0.9618
}\depthtable
\pgfplotstableread{
d  gsmL40S gsmL40Slo gsmL40Shi
2  0.9664  0.9589    0.9892
3  0.9409  0.8932    0.9653
4  0.8995  0.8863    0.9120
5  0.8675  0.8184    0.9166
6  0.7903  0.7255    0.8501
7  0.7223  0.6721    0.8563
}\depthgsmlfortytable

\pgfplotsset{
  b200/.style={color=fwdb200, mark=*, mark size=1.1pt},
  l40s/.style={color=fwdl40s, mark=square*, mark size=1.1pt},
}
\pgfplotsset{
  heb/.style={color=dephumaneval, solid, mark=*, mark size=1.2pt},
  hel/.style={color=dephumaneval, dashed, mark=square*, mark size=1.2pt, mark options={solid, fill=white}},
  gsb/.style={color=depgsm, solid, mark=*, mark size=1.2pt},
  gsl/.style={color=depgsm, dashed, mark=square*, mark size=1.2pt, mark options={solid, fill=white}},
}
\tikzset{
  val/.style={font=\scriptsize\bfseries, inner sep=1pt},
  refnote/.style={font=\scriptsize\itshape, text=black!70, inner sep=1pt},
}

\begin{tikzpicture}
\begin{groupplot}[
  group style={group size=3 by 1, horizontal sep=34pt},
  scale only axis,
  height=0.215\textwidth,
  axis line style={black},
  tick style={black},
  tick align=inside,
  major tick length=3pt,
  tick label style={font=\scriptsize},
  label style={font=\footnotesize},
  title style={font=\small, yshift=-3pt},
  xmajorgrids=true,
  ymajorgrids=true,
  grid style={densely dashed,black!18},
  legend cell align=left,
  legend style={font=\scriptsize, draw=black!30, fill=white, fill opacity=0.9,
    text opacity=1, inner sep=1.5pt, row sep=-1pt},
  every axis plot/.append style={line width=1.1pt},
  clip=false,
]

\nextgroupplot[
    width=0.245\textwidth,
    xmode=log, log basis x=2,
    xmin=360, xmax=3500,
    xtick={400,800,1600,3200}, xticklabels={400,800,1600,3200},
    log ticks with fixed point,
    ymin=0.75, ymax=3.68,
    ytick={1,1.5,2,2.5,3,3.5},
    yticklabel style={/pgf/number format/.cd,fixed,fixed zerofill,precision=1},
    xlabel={Complete sequence length (tokens)},
,
  title={(a) LLaDA-8B-Instruct},
  ylabel={\shortstack{Three-candidate call cost\\(single-call units)}}]
\addplot[black, densely dashed, line width=0.6pt, forget plot] coordinates {(360,3) (3500,3)};
\addplot[black, densely dotted, line width=0.6pt, forget plot] coordinates {(360,1) (3500,1)};
\addplot[b200] table[x=n, y=lladaB200] {\fwdcosttable};
\addplot[l40s] table[x=n, y=lladaL40S] {\fwdcosttable};
\node[val, text=fwdb200, anchor=north west] at (axis cs:415,1.30) {1.33};
\node[val, text=fwdl40s, anchor=north] at (axis cs:450,2.38) {2.44};
\node[val, text=fwdb200, anchor=north east] at (axis cs:3100,2.82) {2.88};
\node[val, text=fwdl40s, anchor=south east] at (axis cs:3100,3.36) {3.30};
\node[refnote, anchor=south east] at (axis cs:3400,1.03) {one single-candidate call};

\nextgroupplot[
    width=0.245\textwidth,
    xmode=log, log basis x=2,
    xmin=360, xmax=3500,
    xtick={400,800,1600,3200}, xticklabels={400,800,1600,3200},
    log ticks with fixed point,
    ymin=0.75, ymax=3.68,
    ytick={1,1.5,2,2.5,3,3.5},
    yticklabel style={/pgf/number format/.cd,fixed,fixed zerofill,precision=1},
    xlabel={Complete sequence length (tokens)},
,
  title={(b) Dream-7B-Instruct}, yticklabels={},
  legend style={at={(0.97,0.06)}, anchor=south east}]
\addplot[black, densely dashed, line width=0.6pt, forget plot] coordinates {(360,3) (3500,3)};
\addplot[black, densely dotted, line width=0.6pt, forget plot] coordinates {(360,1) (3500,1)};
\addplot[b200] table[x=n, y=dreamB200] {\fwdcosttable};
\addlegendentry{B200}
\addplot[l40s] table[x=n, y=dreamL40S] {\fwdcosttable};
\addlegendentry{L40S}
\node[val, text=fwdb200, anchor=north west] at (axis cs:415,1.34) {1.38};
\node[val, text=fwdl40s, anchor=north west] at (axis cs:415,1.98) {2.02};
\node[val, text=fwdb200, anchor=north east] at (axis cs:3100,2.95) {3.01};
\node[val, text=fwdl40s, anchor=south east] at (axis cs:3100,3.27) {3.19};

\nextgroupplot[
  title={(c) Complete-generation speedup},
  width=0.27\textwidth,
  xmin=1.75, xmax=8.25, xtick={2,...,8},
  ymin=0.55, ymax=1.98,
  ytick={0.6,0.8,1.0,1.2,1.4,1.6,1.8},
  yticklabel style={/pgf/number format/.cd,fixed,fixed zerofill,precision=1},
  xlabel={Speculation depth},
  ylabel={Speedup over DirectCTMC},
  ylabel shift=-2pt,
  legend columns=2,
  legend style={at={(0.98,0.97)}, anchor=north east,
    /tikz/every even column/.append style={column sep=4pt}},
]
\addplot[black, densely dashed, line width=0.6pt, forget plot] coordinates {(1.75,1) (8.25,1)};
\addplot[draw=none, name path=helo, forget plot] table[x=d, y=heB200lo] {\depthtable};
\addplot[draw=none, name path=hehi, forget plot] table[x=d, y=heB200hi] {\depthtable};
\addplot[dephumaneval, fill opacity=0.11, forget plot] fill between[of=helo and hehi];
\addplot[draw=none, name path=hllo, forget plot] table[x=d, y=heL40Slo] {\depthtable};
\addplot[draw=none, name path=hlhi, forget plot] table[x=d, y=heL40Shi] {\depthtable};
\addplot[dephumaneval, fill opacity=0.11, forget plot] fill between[of=hllo and hlhi];
\addplot[draw=none, name path=gblo, forget plot] table[x=d, y=gsmB200lo] {\depthtable};
\addplot[draw=none, name path=gbhi, forget plot] table[x=d, y=gsmB200hi] {\depthtable};
\addplot[depgsm, fill opacity=0.11, forget plot] fill between[of=gblo and gbhi];
\addplot[draw=none, name path=gllo, forget plot] table[x=d, y=gsmL40Slo] {\depthgsmlfortytable};
\addplot[draw=none, name path=glhi, forget plot] table[x=d, y=gsmL40Shi] {\depthgsmlfortytable};
\addplot[depgsm, fill opacity=0.11, forget plot] fill between[of=gllo and glhi];
\addplot[heb, forget plot] table[x=d, y=heB200] {\depthtable};
\addplot[hel, forget plot] table[x=d, y=heL40S] {\depthtable};
\addplot[gsb, forget plot] table[x=d, y=gsmB200] {\depthtable};
\addplot[gsl, forget plot] table[x=d, y=gsmL40S] {\depthgsmlfortytable};
\addlegendimage{black!40, solid, line width=1.1pt, mark=*, mark size=1.2pt}
\addlegendentry{B200}
\addlegendimage{black!40, dashed, line width=1.1pt, mark=square*, mark size=1.2pt, mark options={solid, fill=white}}
\addlegendentry{L40S}
\node[font=\scriptsize, text=dephumaneval, anchor=south west, inner sep=1pt] at (axis cs:1.9,1.62) {HumanEval};
\node[font=\scriptsize, text=depgsm, anchor=north west, inner sep=1pt] at (axis cs:1.9,0.84) {GSM8K};
\node[val, text=dephumaneval, anchor=south] at (axis cs:5,1.64) {1.61\(\times\)};
\node[val, text=dephumaneval, anchor=south] at (axis cs:3,1.235) {1.18\(\times\)};
\node[text=depgsm, font=\footnotesize\bfseries, inner sep=0pt] at (axis cs:8,0.60) {\(\times\)};
\node[font=\scriptsize, text=black!70, anchor=east, inner sep=1pt] at (axis cs:7.8,0.60)
  {L40S, GSM8K: out of memory};
\end{groupplot}
\end{tikzpicture}
\endgroup

\caption{
\textbf{Batching cost and complete-generation speedup depend on the hardware.}
(a, b) Forward-only relative cost of one three-candidate target call against one single-candidate call, versus complete sequence length for LLaDA-8B-Instruct and Dream-7B-Instruct on a B200 and an L40S GPU. The dashed line marks the cost of three separate calls and the dotted line one single-candidate call. At short lengths on the B200, one three-state call costs less than three separate calls. As sequence length grows, the three-state call approaches the cost of three separate calls on both GPUs, with the L40S reaching that regime at shorter lengths.
(c) Complete-generation speedup of SpecCTMC over DirectCTMC versus speculation depth for Dream-7B-Instruct on HumanEval and GSM8K with 256 generated tokens, on a B200 (solid) and an L40S (dashed); shaded bands are paired 95\% intervals. The same depth, acceptance, and evaluation reduction give different wall-clock outcomes on the two devices, and depth 8 on GSM8K exceeds L40S memory.}
\label{fig:forward-cost-hardware}
\label{fig:depth-hardware}
\end{figure*}


\begin{table}[t]
\centering
\small
\setlength{\tabcolsep}{4pt}
\caption{\textbf{DirectCTMC vs.\ SpecCTMC on multimodal LLaDA-V.}
NFE is averaged across the fixed evaluation set and sampling seeds.}
\label{tab:nfe-main-lladav}
\begin{tabular}{@{}lrrrrrr@{}}
\toprule
Dataset & $L$ & Depth & {\bf Direct.} & {\bf Spec.} & \shortstack{NFE\\reduction} & Accept. \\
\midrule
MathVerse & 256 & 8 & 256.00 & 43.17 & 5.93$\times$ & 0.698 \\
          & 512 & 8 & 512.00 & 78.04 & 6.56$\times$ & 0.784 \\
\midrule
MathVista & 256 & 8 & 256.00 & 42.41 & 6.04$\times$ & 0.718 \\
          & 512 & 8 & 512.00 & 76.48 & 6.69$\times$ & 0.806 \\
\bottomrule
\end{tabular}
\end{table}

Across the text evaluation in \cref{tab:llada-compact,tab:dream-compact}, SpecCTMC reduces sequential target-model evaluations by factors from 1.729 to 4.379. Acceptance rate is a standard diagnostic for AR speculative decoding literature \citep{wang2026diversed}; it measures the fraction of proposed events accepted without positive-residual correction, while the NFE reduction also depends on how accepted events are grouped into verification rounds. In \Cref{tab:nfe-main-lladav}, on multimodal LLaDA-V, SpecCTMC achieves larger NFE reductions of $5.93$-$6.69\times$ on MathVerse and MathVista, extending these gains to image-conditioned generation.


\paragraph{NFE vs Wall-Clock Time Acceleration.}
The NFE reductions in \cref{tab:llada-compact,tab:dream-compact} count sequential target-model calls. For complete sequence length \(n\) and hidden width \(h\), linear layers flatten the three-state tensor \([3,n,h]\) to \([3n,h]\). Batching combines the states into one model call without removing their matrix computation. \Cref{fig:forward-cost-hardware} shows that, for short sequences on the B200, one three-state call costs less than three separate calls. As sequence length grows, its cost approaches that of three separate calls on both GPUs.

Accordingly, the right panel finds Dream on HumanEval faster at every
B200 depth and at L40S depths 2 to 5, while long-prompt GSM8K is at
parity or slower and exceeds L40S memory at depth 8. This compute-bound
behavior matches \citet{chen2024magicdec}. In the Self Speculative
Decoding control~\citep{gao2025ssd}, approximate caching changes a 1.76
times slowdown into a 1.40 times speedup. Details are given in
\cref{app:nfe-wallclock}.

\subsection{Comparison against approximate samplers}
\label{sec:approximate-samplers}

Approximate solvers reduce sampling cost by advancing several token
updates before recomputing the model's predictions. Tau leaping uses
transition rates evaluated at the beginning of each interval, while the
trapezoidal solver adds an intermediate evaluation to better approximate
how these rates change~\citep{campbell2022continuous,lou2024sedd,ren2025fast}.
Both approaches reduce sequential evaluations through finite-step
approximation. SpecCTMC instead reduces sequential evaluations while
preserving the underlying CTMC distribution. We give the solver
constructions and implementation details in
\cref{app:approximate-solvers}.

\definecolor{specorange}{HTML}{FE8010}
\definecolor{taublue}{HTML}{468FD8}
\definecolor{trapgreen}{HTML}{70AD47}
\definecolor{directgray}{HTML}{404040}

\begin{figure}[t]
\centering
\begingroup
\begin{tikzpicture}
\begin{groupplot}[
  group style={group size=4 by 2,
    horizontal sep=5pt,
    vertical sep=6pt,
    xticklabels at=edge bottom,
    yticklabels at=edge left},
  scale only axis,
  width=0.205\columnwidth,
  height=0.20\columnwidth,
  xmode=log,log basis x=2,
  xmin=0.75,xmax=22,
  xtick={1,4,16},xticklabels={1,4,16},
  ymin=30,ymax=128,
  ytick={50,75,100,125},
  axis line style={black!45},
  tick style={black!45},tick align=inside,
  major tick length=2pt,
  tick label style={font=\fontsize{6}{7}\selectfont},
  every axis x label/.style={at={(0.5,-0.04)},anchor=north},
  label style={font=\scriptsize},
  title style={font=\scriptsize,yshift=-4pt},
  ymajorgrids=true,grid style={black!8},
  clip=false,
  legend columns=4,
  legend style={draw=none,fill=none,font=\fontsize{6}{7}\selectfont,
    /tikz/every even column/.append style={column sep=2pt}},
  legend cell align=left,
]
\nextgroupplot[title={\(K=1\)},ylabel={Mean acc.\ (\%)},legend to name=solverretentionlegend]
\addplot[black!40,densely dotted,forget plot] coordinates {(0.75,100) (22,100)};
\addplot[only marks,color=directgray,mark=square*,mark size=1.5pt] coordinates {(1,100)};
\addlegendentry{DirectCTMC}
\addplot[color=taublue,line width=0.8pt,mark=triangle*,mark size=1.8pt] coordinates {(0.99221790,94.44444444) (1.32124352,61.11111111) (3.92307692,55.55555556) (15.00000000,50.00000000)};
\addlegendentry{\(\tau\)-leaping}
\addplot[color=trapgreen,line width=0.8pt,mark=diamond*,mark size=1.8pt] coordinates {(1.97674419,66.66666667) (15.00000000,38.88888889)};
\addlegendentry{Trapezoidal}
\addplot[only marks,color=specorange,mark=star,mark size=3pt,mark options={line width=0.9pt}] coordinates {(2.80219780,111.11111111)};
\addlegendentry{SpecCTMC (ours)}
\node[anchor=south west,font=\fontsize{5.5}{6}\selectfont,text=black!55,inner sep=1.5pt] at (rel axis cs:0,0) {28.1\%};
\nextgroupplot[title={\(K=2\)}]
\addplot[black!40,densely dotted,forget plot] coordinates {(0.75,100) (22,100)};
\addplot[only marks,color=directgray,mark=square*,mark size=1.5pt] coordinates {(1,100)};
\addplot[color=taublue,line width=0.8pt,mark=triangle*,mark size=1.8pt] coordinates {(0.98832685,115.62500000) (1.31606218,90.62500000) (3.90769231,78.12500000) (14.94117647,75.00000000)};
\addplot[color=trapgreen,line width=0.8pt,mark=diamond*,mark size=1.8pt] coordinates {(1.96899225,96.87500000) (14.94117647,62.50000000)};
\addplot[only marks,color=specorange,mark=star,mark size=3pt,mark options={line width=0.9pt}] coordinates {(2.79120879,109.37500000)};
\node[anchor=south west,font=\fontsize{5.5}{6}\selectfont,text=black!55,inner sep=1.5pt] at (rel axis cs:0,0) {25.0\%};
\nextgroupplot[title={\(K=4\)}]
\addplot[black!40,densely dotted,forget plot] coordinates {(0.75,100) (22,100)};
\addplot[only marks,color=directgray,mark=square*,mark size=1.5pt] coordinates {(1,100)};
\addplot[color=taublue,line width=0.8pt,mark=triangle*,mark size=1.8pt] coordinates {(0.98929961,97.14285714) (1.31735751,82.85714286) (3.91153846,75.71428571) (14.95588235,55.71428571)};
\addplot[color=trapgreen,line width=0.8pt,mark=diamond*,mark size=1.8pt] coordinates {(1.97093023,85.71428571) (14.95588235,54.28571429)};
\addplot[only marks,color=specorange,mark=star,mark size=3pt,mark options={line width=0.9pt}] coordinates {(2.80551724,100.00000000)};
\node[anchor=south west,font=\fontsize{5.5}{6}\selectfont,text=black!55,inner sep=1.5pt] at (rel axis cs:0,0) {27.3\%};
\nextgroupplot[title={\(K=8\)}]
\addplot[black!40,densely dotted,forget plot] coordinates {(0.75,100) (22,100)};
\addplot[only marks,color=directgray,mark=square*,mark size=1.5pt] coordinates {(1,100)};
\addplot[color=taublue,line width=0.8pt,mark=triangle*,mark size=1.8pt] coordinates {(0.98978599,91.78082192) (1.31800518,76.71232877) (3.91346154,76.02739726) (14.96323529,46.57534247)};
\addplot[color=trapgreen,line width=0.8pt,mark=diamond*,mark size=1.8pt] coordinates {(1.97189922,83.56164384) (14.96323529,49.31506849)};
\addplot[only marks,color=specorange,mark=star,mark size=3pt,mark options={line width=0.9pt}] coordinates {(2.81077348,95.89041096)};
\node[anchor=south west,font=\fontsize{5.5}{6}\selectfont,text=black!55,inner sep=1.5pt] at (rel axis cs:0,0) {28.5\%};
\nextgroupplot[ylabel={Pass@\(K\) (\%)}]
\addplot[black!40,densely dotted,forget plot] coordinates {(0.75,100) (22,100)};
\addplot[only marks,color=directgray,mark=square*,mark size=1.5pt] coordinates {(1,100)};
\addplot[color=taublue,line width=0.8pt,mark=triangle*,mark size=1.8pt] coordinates {(0.99221790,94.44444444) (1.32124352,61.11111111) (3.92307692,55.55555556) (15.00000000,50.00000000)};
\addplot[color=trapgreen,line width=0.8pt,mark=diamond*,mark size=1.8pt] coordinates {(1.97674419,66.66666667) (15.00000000,38.88888889)};
\addplot[only marks,color=specorange,mark=star,mark size=3pt,mark options={line width=0.9pt}] coordinates {(2.80219780,111.11111111)};
\node[anchor=south west,font=\fontsize{5.5}{6}\selectfont,text=black!55,inner sep=1.5pt] at (rel axis cs:0,0) {28.1\%};
\nextgroupplot[]
\addplot[black!40,densely dotted,forget plot] coordinates {(0.75,100) (22,100)};
\addplot[only marks,color=directgray,mark=square*,mark size=1.5pt] coordinates {(1,100)};
\addplot[color=taublue,line width=0.8pt,mark=triangle*,mark size=1.8pt] coordinates {(0.98832685,123.07692308) (1.31606218,92.30769231) (3.90769231,76.92307692) (14.94117647,76.92307692)};
\addplot[color=trapgreen,line width=0.8pt,mark=diamond*,mark size=1.8pt] coordinates {(1.96899225,92.30769231) (14.94117647,65.38461538)};
\addplot[only marks,color=specorange,mark=star,mark size=3pt,mark options={line width=0.9pt}] coordinates {(2.79120879,107.69230769)};
\node[anchor=south west,font=\fontsize{5.5}{6}\selectfont,text=black!55,inner sep=1.5pt] at (rel axis cs:0,0) {40.6\%};
\nextgroupplot[]
\addplot[black!40,densely dotted,forget plot] coordinates {(0.75,100) (22,100)};
\addplot[only marks,color=directgray,mark=square*,mark size=1.5pt] coordinates {(1,100)};
\addplot[color=taublue,line width=0.8pt,mark=triangle*,mark size=1.8pt] coordinates {(0.98929961,92.68292683) (1.31735751,75.60975610) (3.91153846,78.04878049) (14.95588235,63.41463415)};
\addplot[color=trapgreen,line width=0.8pt,mark=diamond*,mark size=1.8pt] coordinates {(1.97093023,75.60975610) (14.95588235,56.09756098)};
\addplot[only marks,color=specorange,mark=star,mark size=3pt,mark options={line width=0.9pt}] coordinates {(2.80551724,90.24390244)};
\node[anchor=south west,font=\fontsize{5.5}{6}\selectfont,text=black!55,inner sep=1.5pt] at (rel axis cs:0,0) {64.1\%};
\nextgroupplot[]
\addplot[black!40,densely dotted,forget plot] coordinates {(0.75,100) (22,100)};
\addplot[only marks,color=directgray,mark=square*,mark size=1.5pt] coordinates {(1,100)};
\addplot[color=taublue,line width=0.8pt,mark=triangle*,mark size=1.8pt] coordinates {(0.98978599,93.87755102) (1.31800518,79.59183673) (3.91346154,89.79591837) (14.96323529,71.42857143)};
\addplot[color=trapgreen,line width=0.8pt,mark=diamond*,mark size=1.8pt] coordinates {(1.97189922,75.51020408) (14.96323529,73.46938776)};
\addplot[only marks,color=specorange,mark=star,mark size=3pt,mark options={line width=0.9pt}] coordinates {(2.81077348,93.87755102)};
\node[anchor=south west,font=\fontsize{5.5}{6}\selectfont,text=black!55,inner sep=1.5pt] at (rel axis cs:0,0) {76.6\%};
\end{groupplot}
\node[anchor=south,yshift=2pt] at ($(group c1r1.north west)!0.5!(group c4r1.north east)+(0,8pt)$)
  {\pgfplotslegendfromname{solverretentionlegend}};
\node[anchor=north,font=\scriptsize,yshift=-9pt] at ($(group c1r2.south west)!0.5!(group c4r2.south east)$)
  {Sequential evaluation reduction vs.\ DirectCTMC (\(\times\))};
\end{tikzpicture}
\endgroup
\caption{\textbf{Output quality versus sequential evaluation reduction.}
LLaDA-8B-Instruct, GSM8K, 256 tokens, \(T=1.0\).
Columns: \(K\) answers per question. Rows: mean accuracy and Pass@\(K\),
normalized to DirectCTMC at the same \(K\) (absolute DirectCTMC score in the lower-left corner).
The horizontal axis is the ratio of median cumulative sequential target-model
evaluations (DirectCTMC / method).
From left to right, \(\tau\)-leaping uses 256, 192, 64 and 16 steps and
trapezoidal uses 64 and 8 steps;
at \(S+1\) and \(2S+1\) evaluations per answer, 16-step \(\tau\)-leaping and
8-step trapezoidal are cost matched.}
\label{fig:solver-retention-by-k}
\end{figure}

We compare these methods on GSM8K using LLaDA-8B-Instruct with
256 generated tokens and temperature 1.0.
\Cref{fig:solver-retention-by-k} reports mean answer accuracy and
Pass@\(K\) for \(K\in\{1,2,4,8\}\) independent attempts per question.
Each metric is normalized to DirectCTMC at the same \(K\).
Evaluation reduction uses the corresponding median cumulative
sequential target-model evaluations. We account for the trapezoidal
solver's two evaluations per step; in particular, eight trapezoidal
steps and sixteen tau-leaping steps both require 17 evaluations per
answer.

Coarse solver schedules
substantially reduce evaluations but lower mean answer accuracy.
Additional attempts recover some solution coverage: at \(K=8\), the
coarsest tau-leaping and trapezoidal settings achieve approximately
55\% Pass@8 despite mean answer accuracy of approximately 14\%.
Increasing the step budget does not produce a monotonic improvement
in every observed metric, and trapezoidal sampling does not uniformly
outperform tau leaping at their matched evaluation budget.

SpecCTMC reduces sequential evaluations by approximately
\(2.8\times\) across these answer counts without introducing a solver
approximation. At \(K=8\), SpecCTMC and 256-step tau leaping both achieve
71.88\% Pass@8, using 724 and 2,056 median cumulative evaluations,
respectively. Thus, SpecCTMC reaches the same observed coverage with
approximately \(2.84\times\) fewer sequential evaluations in this
comparison, while preserving the DirectCTMC sampling law.

\begin{figure}[t]
\centering
\begingroup

\definecolor{specorange}{HTML}{FE8010}
\definecolor{taublue}{HTML}{468FD8}
\definecolor{trapgreen}{HTML}{70AD47}
\definecolor{directgray}{HTML}{404040}

\begin{tikzpicture}
\begin{groupplot}[
  group style={
    group size=1 by 2,
    vertical sep=58pt
  },
  scale only axis,
  width=0.82\columnwidth,
  height=0.40\columnwidth,
  xmode=log,
  log basis x=2,
  ymin=28,
  ymax=100,
  ytick={30,40,50,60,70,80,90,100},
  axis line style={black},
  tick style={black},
  tick align=inside,
  major tick length=3pt,
  tick label style={font=\scriptsize},
  label style={font=\footnotesize},
  xlabel style={align=center},
  xmajorgrids=true,
  ymajorgrids=true,
  grid style={densely dashed,black!18},
  minor x tick num=0,
  clip=false,
  legend cell align=left,
  legend style={
    draw=none,
    fill=white,
    fill opacity=0.85,
    text opacity=1,
    font=\scriptsize,
    inner sep=0pt,
    column sep=3pt
  },
]

\nextgroupplot[
  xlabel={Independent answers per question, \(K\)\\[5pt]
    {\small\bfseries (a) Stochastic sampling expands coverage}},
  ylabel={Pass@\(K\) (\%)},
  xmin=0.9,
  xmax=128,
  xtick={1,2,4,8,16,32,64},
  xticklabels={1,2,4,8,16,32,64},
  legend columns=2,
  legend style={at={(0.98,0.03)},anchor=south east},
]

\addplot[
  black!75,
  dashed,
  line width=0.8pt,
  forget plot
] coordinates {
  (0.9,81.25)
  (128,81.25)
};

\node[
  anchor=south west,
  font=\scriptsize,
  fill=white,
  inner sep=1pt,
  text=black!75
] at (axis cs:1.02,83.0) {
  \shortstack[l]{Best matched deterministic\\81.25\%}
};

\addplot[
  color=taublue!30!white,
  line width=0.8pt,
  mark=*,
  mark size=1.5pt
] coordinates {
  (1,48.44)
  (2,57.81)
  (4,75.00)
  (8,82.81)
  (16,85.94)
  (32,93.75)
  (64,96.88)
};
\addlegendentry{\(T=0.1\)}

\addplot[
  color=taublue!55!white,
  line width=0.8pt,
  mark=*,
  mark size=1.5pt
] coordinates {
  (1,46.88)
  (2,56.25)
  (4,65.63)
  (8,79.69)
  (16,85.94)
  (32,95.31)
  (64,96.88)
};
\addlegendentry{\(T=0.3\)}

\addplot[
  color=taublue!78!white,
  line width=1.5pt,
  mark=*,
  mark size=2.1pt
] coordinates {
  (1,45.31)
  (2,54.69)
  (4,73.44)
  (8,79.69)
  (16,89.06)
  (32,93.75)
  (64,96.88)
};
\addlegendentry{\(T=0.5\)}

\addplot[
  color=taublue!100!white,
  line width=0.8pt,
  mark=*,
  mark size=1.5pt
] coordinates {
  (1,31.25)
  (2,43.75)
  (4,57.81)
  (8,71.88)
  (16,81.25)
  (32,85.94)
  (64,95.31)
};
\addlegendentry{\(T=1.0\)}

\draw[
  Latex-Latex,
  color=taublue,
  line width=0.8pt
]
  (axis cs:96,81.25) --
  (axis cs:96,96.88);

\node[
  anchor=north east,
  font=\scriptsize\bfseries,
  text=taublue,
  fill=white,
  inner sep=1pt
] at (axis cs:104,80.5) {\(+15.63\) pp};

\nextgroupplot[
  xlabel={Cumulative sequential target-model evaluations\\[5pt]
    {\small\bfseries (b) Exact speculation reduces sampling cost}},
  xmin=64,
  xmax=21000,
  xtick={128,512,2048,8192},
  xticklabels={128,512,{2,048},{8,192}},
  ylabel={Pass@\(K\) (\%)},
  legend columns=1,
  legend style={at={(0.02,0.97)},anchor=north west},
]

\addplot[
  black!75,
  dashed,
  line width=0.8pt,
  forget plot
] coordinates {
  (64,81.25)
  (21000,81.25)
};

\addplot[
  color=directgray,
  line width=1pt,
  mark=square*,
  mark size=2pt
] coordinates {
  (255,28.12)
  (508,40.63)
  (1017,64.06)
  (2035,76.56)
  (4071,79.69)
  (8140,85.94)
  (16281.5,93.75)
};
\addlegendentry{DirectCTMC}

\addplot[
  color=specorange,
  line width=1.35pt,
  mark=star,
  mark size=3.7pt,
  mark options={solid,line width=0.8pt}
] coordinates {
  (91,31.25)
  (182,43.75)
  (362.5,57.81)
  (724,71.88)
  (1450,81.25)
  (2901,85.94)
  (5800,95.31)
};
\addlegendentry{SpecCTMC}

\draw[
  Latex-Latex,
  color=specorange,
  line width=0.8pt,
  shorten <=4pt,
  shorten >=4pt
]
  (axis cs:2901,85.94) --
  (axis cs:8140,85.94);

\node[
  anchor=south east,
  align=center,
  font=\scriptsize,
  text=specorange,
  fill=white,
  inner sep=1pt
] at (rel axis cs:0.98,0.5) {
  \textbf{\(2.81\times\) fewer eval}\\
  at \(K=32\)
};

\end{groupplot}
\end{tikzpicture}
\endgroup

\caption{
{\bf Stochastic sampling expands solution coverage, and exact speculation reduces its sequential cost.}
All results use LLaDA-8B-Instruct on five-shot GSM8K with 256 generated tokens. The dashed line is the strongest matched deterministic configuration, Fast-dLLM factor \(f=1.0\) (81.25\%; \cref{tab:deterministic-matched}). (a): empirical Pass@\(K\) for four selected fixed token temperatures; \(T=0.5\) is highlighted. (b): measured DirectCTMC and SpecCTMC results at \(T=1.0\), with points corresponding to \(K\in\{1,2,4,8,16,32,64\}\). At \(K=32\), both attain 85.94\% observed coverage, while median cumulative sequential evaluations decrease from 8,140 to 2,901 (\(2.81\times\)).
}
\label{fig:stochastic-coverage}
\end{figure}

\subsection{Solution Coverage from Repeated Stochastic Sampling}

Deterministic greedy denoising is effective when only one answer is required. Additional attempts do not expand its solution coverage because the decoder returns the same output each time. Stochastic sampling explores alternative outputs, so additional computation can discover solutions missed by the single deterministic decoding trajectory.

Token temperature controls the token sampled at a selected position, while position temperature controls the order in which masked positions are unmasked. Mean accuracy over \(K\) attempts averages correctness across all generated answers through attempt \(K\). Pass@\(K\) is the percentage of evaluation problems solved at least once. Variation in unmasking order can restore Pass@\(K\) scaling and improve coverage at a matched evaluation budget~\citep{olausson2026twotemperatures}. The temperature that gives the best coverage can also change with the number of attempts~\citep{lamont2026freelunch}. These results motivate exact acceleration of stochastic sampling. Approximate acceleration can change the chosen sampling distribution. SpecCTMC reduces sequential target-model evaluations while preserving the complete CTMC sampling law at each fixed temperature.


We additionally evaluated recent deterministic decoding methods
to establish a strong reference for this comparison.
Greedy denoising and the confidence-threshold and factor decoders
from Fast-dLLM~\citep{wu2025fastdllm} were run with the same checkpoint revision, prompts, evaluation
questions, generation length, scorer, and timing procedure.
Their accuracies range from 78.13\% to 81.25\%, with the factor
decoder providing the strongest matched deterministic result.
Under the tested configurations, repeated runs produce the same
answer, so their solution coverage remains constant as the
number of attempts increases.
Pass@\(K\) measures solution coverage across these candidate answers.
These experiments do not evaluate a selector that chooses one final answer.
We therefore use 81.25\% as the deterministic reference in
\cref{fig:stochastic-coverage}.
Complete results, including additional greedy-derived speculative
methods such as PSD, trajectory-level speculative decoding,
and Spiffy~\citep{sun2026psd,pan2026trajectory,agrawal2026spiffy},
are reported in \cref{app:deterministic-references}.

Lower token temperatures improve single-answer accuracy, while repeated
stochastic answers increase solution coverage. The highlighted \(T=0.5\)
curve exceeds the strongest matched deterministic result by \(K=16\),
and all four displayed temperatures do so by \(K=32\). At \(T=1.0\) and
\(K=32\), SpecCTMC matches DirectCTMC's coverage with \(2.81\times\)
fewer median cumulative sequential evaluations while preserving the
CTMC law.

We separate the effects of token choice and unmasking order in
\cref{app:token-choice-reveal-order}.

\subsection{Composition of SpecCTMC with complementary improvements}
\label{sec:composition}

The practical goal for diffusion language models is to match the quality
of autoregressive models while reducing the sequential work required to
generate an answer. Autoregressive inference benefits from key/value
caching, but a 256-token answer still requires approximately 256
dependent decoding steps~\citep{leviathan2023fast,chen2023speculative}. SpecCTMC and caching address complementary
parts of this cost: SpecCTMC shortens the sequence of dependent target
evaluations, while caching reduces the work performed in each
evaluation.

We first combine depth-three SpecCTMC with a prompt-only key/value cache
adapted from Fast-dLLM~\citep{wu2025fastdllm}. The cache stores the fixed
prompt, recomputes the generated positions, and refreshes the prompt
representation after a fixed number of unmasked tokens. On GSM8K
with LLaDA-8B-Instruct and 256 generated tokens, refresh
intervals of 8 and 16 make SpecCTMC 1.70 and 1.94 times faster than the
corresponding cached sequential CTMC. The two settings reduce sequential
target evaluations by factors of 2.53 and 2.59. Counting each cache refresh
as an additional sequential model evaluation gives 133.5 and 115.3
evaluations, respectively, compared with approximately 256 sequential
model evaluations for standard autoregressive decoding. At both settings, the 95 percent interval for the cached sequential method's accuracy difference from uncached DirectCTMC contains zero.
SpecCTMC preserves the corresponding cached sequential process. This
process approximates complete prompt-and-answer recomputation because it
refreshes the prompt representation only at fixed intervals.

We next test whether the same mechanism transfers to the Fast-dLLM-v2
checkpoint released with DualDiffusion~\citep{goyal2026dualdiffusion}.
Predictions from a previous call to this checkpoint propose up to four future events. We verify each proposal using probabilities evaluated at its tentative state and correct the first rejected proposal. On GSM8K with 32 generated tokens, this reduces sequential checkpoint evaluations from 32.0 to
10.25, or 3.12 times. Including 1.5 proposal evaluations gives 11.75
checkpoint evaluations per answer. The exactness tests pass, while
median generation time remains 55.63 seconds for the event-by-event
sampler and 55.70 seconds for SpecCTMC. This experiment shows that SpecCTMC can use Fast-dLLM-v2 checkpoint probabilities for both proposal and exact verification. 
It does not test composition with the native Fast-dLLM-v2 cache or confidence-based commitment rule. 
Detailed results appear in \cref{app:composition}.

\section{Limitations and Conclusion}
\label{sec:conclusion}

SpecCTMC reduces sequential target-model evaluations for the natural-event CTMC induced by an unchanged masked denoiser, while exact verification and positive-residual correction preserve DirectCTMC's event-time and mark law. Our experiments on LLaDA and Dream show lower latency across most measured conditions without a separate draft model, and the cache experiment shows that SpecCTMC composes with cache reuse. The method complements other exact CTMC simulators. Uniformization replaces state-dependent event rates with a dominating Poisson clock and virtual self-transitions, and has been developed for reverse discrete diffusion without time-discretization error~\citep{chen2024uniformization}. A matched comparison can determine when its virtual-event cost is preferable to SpecCTMC's additional target rows.

For SpecCTMC, elapsed time depends on sequence length, memory, and the target model's batch-cost curve because rows after the first rejection may be unused~\citep{chen2024magicdec,ho2026specfold}; our hardware study quantifies these effects in \cref{app:nfe-wallclock}. The influence condition in \cref{thm:influence-complexity} gives a sufficient condition for long accepted prefixes and linear expected target-row work~\citep{liu2024parallelize}. Larger changes in the remaining token predictions can reduce acceleration, but they do not affect the path-law guarantee in \cref{thm:marked-path}. These results establish exact speculative batching as a practical way to exchange parallel target work for lower sequential depth while preserving the chosen stochastic decoding process.

\clearpage

\bibliography{references}
\bibliographystyle{mlsys2025}

\section*{Impact Statement}

This work develops an inference algorithm for sampling from a fixed diffusion language model more efficiently under explicit mathematical conditions. Faster generation can reduce latency and accelerator use for applications that already deploy these models. It can also lower the cost of producing low-quality, misleading, or harmful text. The method does not change model weights, training data, or safety behavior, so it neither removes nor introduces content safeguards by itself. Deployments remain responsible for model selection, access controls, output evaluation, and monitoring. Our restricted code evaluation executes generated programs under process, memory, time, import, file, and network limits rather than treating unrestricted execution as a benchmark convenience.

\section*{AI Usage Disclosure}

This section describes how generative AI and AI-based tools were used in producing this work. The authors take full responsibility for all content, have checked every AI-assisted output against the underlying data and derivations, and made every scientific claim and decision themselves.

\textbf{Tools.} We used ChatGPT Work and Claude Code.

\textbf{Ideation and method design.} The ideation and method design comes entirely from the authors. No help comes from AI tools in this stage.

\textbf{Literature search.} AI tools helped retrieve related papers and organize comparisons. The authors read the primary sources and selected the citations and claims.

\textbf{Experiment design and code implementation.} We used closed-source Claude agent models to scaffold experiment-evaluation and analysis scripts and to review code.

\textbf{Figures, tables and writing.} AI tools helped prepare figure and table templates and edit manuscript prose from author instructions and verified results. The authors checked the resulting artifacts and text against the underlying evidence.

\clearpage
\appendix

\makeatletter
\let\addcontentsline\standardaddcontentsline
\makeatother
\section*{Appendix}
\begingroup
\small
\noindent\textbf{Contents}\par
\makeatletter
\renewcommand*\l@section{\@dottedtocline{1}{0em}{2.3em}}
\renewcommand*\l@subsection{\@dottedtocline{2}{1.5em}{3.2em}}
\@starttoc{toc}
\makeatother
\endgroup

\crefname{appendix}{appendix}{appendices}
\Crefname{appendix}{Appendix}{Appendices}
\crefalias{section}{appendix}
\crefalias{subsection}{appendix}
\crefalias{subsubsection}{appendix}

\section{Terminology and Notations}
\label{app:terminology}

We collect terms and symbols used throughout the main paper and the appendix, shown in \Cref{tab:terminology}.

\begin{table*}[t]
\centering
\small
\setlength{\tabcolsep}{5pt}
\renewcommand{\arraystretch}{1.08}
\caption{Terminology and notation used throughout the main paper and appendix.}
\label{tab:terminology}
\begin{tabular}{>{\raggedright\arraybackslash}p{0.25\textwidth}>{\raggedright\arraybackslash}p{0.61\textwidth}}
\toprule
Term or symbol & Meaning \\
\midrule
\multicolumn{2}{@{}l}{\textit{Process}} \\
DirectCTMC & Event-by-event exact simulation of the absorbing-mask CTMC: one target-model evaluation per unmasking event. \\
SpecCTMC & Exact speculative simulation of the same CTMC. \\
Natural event & One masked position unmasked at a random event time. \\
Mark \(z=(d,v)\) & Position \(d\) and replacement token \(v\) of one event. \\
State update \(\Phi(x,(d,v))\) & Sequence equal to \(x\) except that position \(d\) is replaced by token \(v\). \\
Product distribution \(Q_a^{\otimes r}\) & Distribution of \(r\) independent binary pairs: \(Q_a^{\otimes r}(x)=\prod_{j=1}^{r}Q_a(x_{2j-1},x_{2j})\). \\
Masked positions \(A(x)\), count \(m(x)\) & Positions in state \(x\) equal to the mask token, and their number. \\
Target kernel \(P_{t,x}\) & Conditional mark distribution at time \(t\) and state \(x\): uniform position, model token probability. \\
Exit intensity \(\lambda_t(x)=m(x)/t\) & Total event rate; depends only on time and mask count. \\
Cutoff \(\varepsilon\), deterministic completion & Simulation stops below model time \(\varepsilon\); remaining masks are then filled using a fixed rule. \\
Event-time sequence \(\mathcal C=(N,T_1,\ldots,T_N)\) & Number and times of events at or above the cutoff, sampled before their marks. \\
Greedy denoising & Deterministic finite-step decoder: confidence-ordered unmasking and argmax token. \\
Tau leaping, trapezoidal solver & Fixed-step approximate CTMC solvers of first and second order.  \\
\midrule
\multicolumn{2}{@{}l}{\textit{Speculation}} \\
Speculation depth \(K\) & Maximum number of events drafted per round; \(d\) denotes a selected depth.  \\
Round, draft length \(L_r\) & One draft-evaluate-verify cycle and its number of drafted events.  \\
Proposal \(D_i\), draft \(\widehat Z_i\), \(\widehat X_i\) & Proposal distribution, proposed mark, and tentative state at draft index \(i\).  \\
Cached target proposal & Proposal that initially uses probabilities from an initial target-model evaluation and subsequently uses target probabilities evaluated at the most recently verified tentative state.  \\
Target row \(P_i\) & Target kernel evaluated at tentative pre-update state \(\widehat X_{i-1}\); target rows are evaluated in batches.  \\
Acceptance \(a_i(z)\), \(\alpha_i\) & Maximal-coupling acceptance probability of mark \(z\) and its average over \(D_i\).  \\
Residual \(R_i\) & Positive-part correction distribution sampled at the first rejection. \\
Accepted prefix \(A_r\), committed \(C_r\) & Accepted events in round \(r\) and committed events including the corrected one. \\
Logical target rows \(W\), charged rows \(\widetilde W\), evaluated states \(W_{\mathrm{eval}}\) & Work counts used in the analysis; rows used for initialization \(B\), deterministic completion \(F\), and padding are counted separately. \\
Row consistency & Each batched target row equals the kernel of its own input state (Condition C6).  \\
\midrule
\multicolumn{2}{@{}l}{\textit{Evaluation Metrics}} \\
NFE, sequential target-model evaluation & Target evaluation that must finish before the next state-dependent evaluation begins. \\
NFE reduction \(R_{\mathrm{NFE}}=N_D/N_S\) & Mean DirectCTMC NFE divided by mean SpecCTMC NFE. \\
Acceptance rate & Fraction of drafted events accepted during verification. \\
Wall-clock ratio \(T_D/T_S\) & DirectCTMC total generation time divided by SpecCTMC total generation time; faster, parity, or slower by the paired 95\% interval. \\
Per-evaluation cost factor \(E\) & \(E=(T_D/N_D)/(T_S/N_S)\), so \(T_D/T_S=R_{\mathrm{NFE}}E\). \\
Call cost \(C_d(n)\) & Time of one target call with \(d\) states at complete sequence length \(n\). \\
Complete sequence length & Prompt plus generated tokens processed by each evaluation. \\
Target batch & Number of states per physical target call, specified by the experiment.  \\
Mean accuracy at \(K\), Pass@\(K\) & Average correctness over \(K\) attempts; fraction of problems solved at least once in \(K\) attempts.  \\
Accuracy ratio & Method accuracy divided by DirectCTMC accuracy. \\
\bottomrule
\end{tabular}
\end{table*}
\FloatBarrier
\section{Assumptions, Complete Proofs and Edge Cases}
\label{app:proofs}

This section states the hypotheses for \cref{thm:marked-path} and gives
its complete proof.  The hypotheses separate the probabilistic argument
from the implementation properties that must be checked for a concrete
target-model evaluator.

\subsection{Assumptions for exactness}
\label{app:exactness-assumptions}

Let \(\mathcal Z\) be a fixed finite ambient mark space.  At state \(x\),
the valid marks form a subset \(\mathcal Z(x)\subseteq\mathcal Z\), and
kernels on valid marks are extended by zero to \(\mathcal Z\).
\Cref{thm:marked-path} uses the following conditions.

\begin{enumerate}[label=C\arabic*., leftmargin=2em]
  \item \label{cond:initial}
  DirectCTMC and SpecCTMC use the same initial-state law and the same
  cutoff \(\varepsilon\).  Their number of events at or above the cutoff is finite almost
  surely.

  \item \label{cond:clock}
  Conditional on the initial state and each committed history, both
  methods use a common clock law that remains valid under every mark
  history produced by correction.  The random primitives used to draw
  this clock are independent of the proposal, verification, and
  correction primitives.  For the absorbing-mask generator,
  \cref{lem:retained-clock} verifies this condition.

  \item \label{cond:target}
  At every sampled event time \(t\) and state \(x\) with positive exit
  intensity, \(P_{t,x}\) is a normalized probability mass function on
  \(\mathcal Z(x)\).  No natural-event mark is requested at zero exit
  intensity.

  \item \label{cond:draft}
  Each draft kernel \(D_{r,i}\) is selected causally from information
  available before proposal \(i\) of round \(r\).  The proposal
  \(\widehat Z_{r,i}\sim D_{r,i}\) uses a fresh draw.  Its verification
  uniform is independent of that draw conditional on the available
  information.  Following rejection, the correction uses a fresh draw
  from the normalized positive part of the full joint mark residual
  \([P_{r,i}-D_{r,i}]_+\) on \(\mathcal Z\).

  \item \label{cond:stopping}
  Verification proceeds in draft order.  It commits a correction at the
  first rejection and discards the complete later tentative suffix.
  Decisions at an earlier index do not depend on that later suffix.

  \item \label{cond:rows}
  Every target row used by verification equals the declared kernel for
  that row's own time and tentative input state, independently of the other
  batch rows and any padding rows.  Outputs of padding rows are
  discarded.

  \item \label{cond:depth}
  The depth of a round is selected from pre-round information before the
  proposal and verification variables of that round are drawn.

  \item \label{cond:update}
  Both methods use the same deterministic state update
  \(X_j=\Phi(X_{j-1},Z_j)\), the same cutoff rule, and the same deterministic completion rule at the cutoff, including all tie
  rules.
\end{enumerate}

Conditions C1 to C8 are hypotheses of the mathematical theorem.  In the
implementation, the target and draft rows are normalized over tokens
other than the mask token; event, proposal, verification, and residual draws use separate
random-number generators; verification stops at the first rejection;
and the fixed-size target evaluator duplicates only padding rows and
discards their outputs.  The source also checks that every reached draft
input state equals the committed state.  These checks establish the stated
control flow.  Condition C6 additionally declares which finite-precision
target operator the theorem treats as its kernel; the released-checkpoint
tests check representative rows rather than proving this numerical
identity for every possible input.

\subsection{Masking schedule and time coordinate}
\label{app:masking-schedule}

Let \(\rho(t)\) be the probability that an answer position is masked at
model time \(t\).  Assume that \(\rho\) is differentiable and strictly
increasing, and that all answer positions use this absorbing-mask
schedule.  Conditional on a position being masked at time \(t\), its
probability of becoming unmasked as time decreases to \(t-h\) is
\begin{equation}
  \frac{\rho(t)-\rho(t-h)}{\rho(t)}
  =\frac{\rho'(t)}{\rho(t)}h+o(h).
\end{equation}
The reverse rate for replacing masked position \(d\) by replacement token \(v\)
is therefore
\begin{equation}
  q_t(x,(d,v))
  =\frac{\rho'(t)}{\rho(t)}
   p_\theta(v\mid x,d,t).
  \label{eq:general-schedule-rate}
\end{equation}
If \(m(x)\) answer positions remain masked, the probabilities over
replacement tokens sum to one, so
\begin{equation}
  \lambda_t(x)
  =m(x)\frac{\rho'(t)}{\rho(t)}.
  \label{eq:general-schedule-total-rate}
\end{equation}
Thus, the total exit intensity depends on model time and the number of
remaining masks, but not on the position or token selected by an event.
For \(E\sim\operatorname{Exp}(1)\), inversion of the accumulated exit
intensity gives
\begin{equation}
  T_{\mathrm{next}}
  =\rho^{-1}\!\left(
     \rho(t)\exp\!\left\{-\frac{E}{m(x)}\right\}
   \right).
  \label{eq:general-schedule-event-time}
\end{equation}

Equivalently, use the masking probability itself as the time coordinate,
\(u=\rho(t)\).  In this coordinate, \cref{eq:general-schedule-rate}
becomes
\begin{equation}
  \widetilde q_u(x,(d,v))
  =\frac{1}{u}
   p_\theta\!\left(v\mid x,d,\rho^{-1}(u)\right).
  \label{eq:mask-probability-coordinate}
\end{equation}
A time-conditioned denoiser must therefore receive the corresponding
original model time \(\rho^{-1}(u)\).  LLaDA, Dream, and LLaDA-V use
\(\rho(t)=t\) for generated-answer tokens~\citep{nie2025llada,ye2025dream,you2025lladav}.
The coordinate change is the identity for the models evaluated here, so
\cref{eq:general-schedule-rate,eq:general-schedule-event-time} reduce to
\cref{eq:generator,eq:event-clock}.  The calculation also shows that the
common-clock argument in Condition C2 applies to any shared schedule
satisfying the assumptions above: replacing a proposed position-token
mark does not change the conditional distribution of later event times
because every valid event removes exactly one mask.

\subsection{Event times sampled before mark generation}

\begin{proof}[Proof of \cref{lem:retained-clock}]
Fix \(X_0=x_0\), let \(m_0=m(x_0)\), and set \(T_0=1\).  For
\(j=1,\ldots,m_0\), define
\begin{equation}
\begin{aligned}
  m_j&=m_0-j+1,
  &E_j&\overset{\mathrm{iid}}{\sim}\operatorname{Exp}(1),\\
  T_j&=T_{j-1}\exp(-E_j/m_j).
\end{aligned}
\end{equation}
For \(0<u<t\), conditioning on \(T_{j-1}=t\) gives
\begin{equation}
\begin{aligned}
  \Pr(T_j\leq u\mid T_{j-1}=t,X_0=x_0)
  &=\Pr\!\left(E_j\geq m_j\log(t/u)\right)\\
  &=\exp\{-m_j\log(t/u)\}\\
  &=\exp\!\left\{-\int_u^t\frac{m_j}{s}\,ds\right\}.
\end{aligned}
\label{eq:clock-survival}
\end{equation}
Because model time decreases, the event \(T_j\leq u\) means that no
jump occurs while the process traverses the interval from \(t\) to
\(u\).  The final expression in \cref{eq:clock-survival} is therefore
the survival probability associated with DirectCTMC's exit intensity
\(m_j/s\).

Every valid natural mark replaces exactly one mask.  After \(j-1\)
events, the mask count is consequently \(m_j\), irrespective of the
positions and token values in the preceding marks.  Thus the conditional
kernel of \(T_j\) depends on the preceding clock only through
\(T_{j-1}\) and the event index \(j\).  Induction over \(j\) proves that
the complete sequence \((T_1,\ldots,T_{m_0})\) has the DirectCTMC clock
law for DirectCTMC and SpecCTMC.

DirectCTMC and SpecCTMC use exactly the event indices
\begin{equation}
  \mathcal I_\varepsilon
  =\{j\in\{1,\ldots,m_0\}:T_j\geq\varepsilon\},
  \qquad N=|\mathcal I_\varepsilon|,
\end{equation}
and therefore have the same law for
\(\mathcal C=(N,T_1,\ldots,T_N)\).  Replacing a proposed mark by any
other valid mark still removes one mask, so it leaves every subsequent
mask-count-dependent clock kernel unchanged.

This argument does not make event times independent of marks.  The mark
kernel \(P_{T_j,X_{j-1}}\) may depend on the realized time \(T_j\).
It establishes only that the clock primitives can be drawn independently
and that their conditional clock law remains valid after mark
correction.
\end{proof}

\subsection{Conditional one-event verification}

\begin{lemma}[Conditional maximal-coupling identity]
\label{lem:conditional-maximal-coupling}
Let \(\mathcal G\) be a sigma-field, and let \(P\) and \(D\) be
\(\mathcal G\)-measurable probability mass functions on the finite set
\(\mathcal Z\).  Conditional on \(\mathcal G\), draw
\(\widehat Z\sim D\) and an independent \(U\sim\operatorname{Unif}(0,1)\).
For \(D(z)>0\), set
\begin{equation}
\begin{aligned}
  a(z)&=\min\{1,P(z)/D(z)\},\\
  \beta&=1-\sum_{y\in\mathcal Z}\min\{P(y),D(y)\}.
\end{aligned}
\end{equation}
Accept when \(U\leq a(\widehat Z)\).  If rejection occurs and
\(\beta>0\), use a fresh conditionally independent primitive to draw the
output from
\begin{equation}
  R(z)=\frac{[P(z)-D(z)]_+}{\beta}.
\end{equation}
Then the output \(Z\) has conditional mass function \(P\), and
\begin{equation}
  \Pr(\text{rejection}\mid\mathcal G)
  =\operatorname{TV}(P,D).
\end{equation}
\end{lemma}

\begin{proof}
Since \(P\) and \(D\) are probability mass functions,
\begin{equation}
\begin{aligned}
  \sum_z[P(z)-D(z)]_+
  &=\sum_z\bigl(P(z)-\min\{P(z),D(z)\}\bigr)\\
  &=1-\sum_z\min\{P(z),D(z)\}=\beta.
\end{aligned}
\end{equation}
The equality of the total positive and negative parts of \(P-D\) also
gives
\begin{equation}
  \beta=\frac12\sum_z|P(z)-D(z)|
  =\operatorname{TV}(P,D),
\end{equation}
so \(R\) is normalized whenever \(\beta>0\).

For each \(z\) with \(D(z)>0\), the accepted contribution is
\begin{equation}
\begin{aligned}
  &\Pr(Z=z,\text{ acceptance}\mid\mathcal G)\\
  &\qquad=D(z)a(z)=\min\{D(z),P(z)\}.
\end{aligned}
\end{equation}
The rejected contribution is
\begin{equation}
\begin{aligned}
  &\Pr(Z=z,\text{ rejection}\mid\mathcal G)\\
  &\qquad=\beta R(z)=[P(z)-D(z)]_+.
\end{aligned}
\end{equation}
Their sum equals \(P(z)\).  If \(D(z)=0\), the mark is never proposed;
the rejection branch contributes its target probability because
\([P(z)-D(z)]_+=P(z)\).  Hence
\begin{equation}
  \Pr(Z=z\mid\mathcal G)=P(z)
\end{equation}
for every \(z\in\mathcal Z\).  If \(\beta=0\), the positive part of
\(P-D\) vanishes.  Equal total masses then imply \(P=D\), every proposal
is accepted almost surely, and no residual distribution is needed.
\end{proof}

This is the standard one-step maximal-coupling construction used in
speculative sampling~\citep{debortoli2025accelerated}; it is not a new
coupling identity. A state with no valid mark or zero exit intensity
produces no event and does not enter this verification step.

\subsection{Complete marked-path law}

\begin{proof}[Proof of \cref{thm:marked-path}]
\textbf{Step 1: conditioning on the common clock.}
By Conditions C1 and C2, it is enough to work under a regular
conditional law given
\begin{equation}
  X_0=x_0,
  \qquad
  \mathcal C=c=(n,t_1,\ldots,t_n).
\label{eq:conditioned-clock}
\end{equation}
The conditioning in \cref{eq:conditioned-clock} includes both the
initial state and the random number of events at or above the cutoff.  The case \(n=0\)
will use the empty product.

\textbf{Step 2: causal information.}
Index speculative rounds by \(r\), and let \(n_r\) be the number of
marks committed before round \(r\).  Let \(L_r\) be its depth after
clipping to the remaining sampled event times.  Define \(\mathcal F_r\) to
contain \(X_0\), the complete clock \(\mathcal C\), all pre-round
choices, and all information from completed rounds.  For local index
\(i\), set
\begin{equation}
  \mathcal G_{r,i}
  =\mathcal F_r\vee
  \sigma(\widehat Z_{r,1:i-1},U_{r,1:i-1}).
\label{eq:causal-information}
\end{equation}
The event that verification reaches index \(i\) is
\(\mathcal G_{r,i}\)-measurable.  Condition C4 and the fresh proposal
primitive give
\begin{equation}
  \widehat Z_{r,i}\mid\mathcal G_{r,i}\sim D_{r,i}.
\label{eq:causal-proposal-law}
\end{equation}

The implementation constructs the complete draft and computes its
target rows before verification.  This does not change the exposure in
\cref{eq:causal-information}: each draft kernel depends only on the
preceding draft prefix, each target row depends on its own tentative
input state, and Condition C5 prevents an earlier acceptance decision from
using the later tentative suffix.  The proof may therefore expose the
same stored variables in causal order.  Conditioning on the complete
realized draft would instead condition on \(\widehat Z_{r,i}\) and would make
\cref{eq:causal-proposal-law} false.

\textbf{Step 3: the target row at a reached index.}
Suppose local index \(i\) produces global event \(j\).  The
first \(i-1\) proposals in the round were then accepted.  The tentative input state and the target row consequently satisfy
\begin{equation}
  \widehat X_{r,i-1}=X_{j-1},
  \qquad
  P_{r,i}=P_{t_j,X_{j-1}}
\label{eq:reached-target-row}
\end{equation}
by Conditions C5, C6, and C8.  Applying
\cref{lem:conditional-maximal-coupling} conditional on
\(\mathcal G_{r,i}\) shows that the mark committed at this reached
index has conditional kernel \(P_{t_j,X_{j-1}}\).

The round and local index that produce event \(j\) are random.  Let
\(Y_{r,i}\) denote the mark produced at local index \(i\), namely the
accepted proposal or the residual correction; define it arbitrarily on
the event that verification does not reach \(i\).  For \(r\geq0\) and
\(i\geq1\), define
\begin{equation}
\begin{aligned}
  E_{r,i}^{j}
  &=\{n_r+i=j,\ L_r\geq i\}\\
  &\quad\cap
   \{\widehat Z_{r,1:i-1}\text{ were accepted}\}.
\end{aligned}
\label{eq:global-index-partition}
\end{equation}
For fixed \(j\), the nonempty events in
\cref{eq:global-index-partition} are disjoint and partition the event
that global mark \(j\) is produced.  Let
\begin{equation}
  \mathcal H_j=\sigma(X_0,\mathcal C,Z_1,\ldots,Z_{j-1}).
\end{equation}
On \(E_{r,i}^{j}\), the preceding global history is contained in
\(\mathcal G_{r,i}\), and \cref{eq:reached-target-row} is
\(\mathcal H_j\)-measurable.  The tower property applied on each
partition event gives
\begin{equation}
\begin{aligned}
 &\Pr_{\mathrm{Spec}}(Z_j=z\mid\mathcal H_j)\\
 &\quad=\sum_{r,i}
   \E\!\left[
     \mathbf 1_{E_{r,i}^{j}}
     \mathbf 1_{\{Y_{r,i}=z\}}
     \,\middle|\,\mathcal H_j
   \right]\\
 &\quad=\sum_{r,i}
   \E\!\left[
     \mathbf 1_{E_{r,i}^{j}}
     \Pr(Y_{r,i}=z\mid\mathcal G_{r,i})
     \,\middle|\,\mathcal H_j
   \right]\\
 &\quad=\sum_{r,i}
   \E\!\left[
     \mathbf 1_{E_{r,i}^{j}}
     P_{t_j,X_{j-1}}(z)
     \,\middle|\,\mathcal H_j
   \right]\\
 &\quad=P_{t_j,X_{j-1}}(z).
\end{aligned}
\label{eq:global-conditional-kernel}
\end{equation}
Thus the random conversion from round-local indices to global indices
does not change the target conditional kernel.

\textbf{Step 4: conditional chain rule.}
For a valid mark sequence \(z_{1:n}\), define
\(x_j=\Phi(x_{j-1},z_j)\).  Extending every kernel by zero outside its
valid mark set, repeated application of
\cref{eq:global-conditional-kernel} yields
\begin{equation}
\begin{aligned}
 &\Pr_{\mathrm{Spec}}
   (Z_{1:n}=z_{1:n}\mid X_0=x_0,\mathcal C=c)\\
 &\qquad=\prod_{j=1}^{n}P_{t_j,x_{j-1}}(z_j).
\end{aligned}
\label{eq:clock-conditioned-product}
\end{equation}
For \(n=0\), the product in
\cref{eq:clock-conditioned-product} equals one.  The right side is
exactly DirectCTMC's conditional mark law.

\textbf{Step 5: joint law and deterministic completion.}
Integrating \cref{eq:clock-conditioned-product} against the common law
of \((X_0,\mathcal C)\) proves equality of the event count, sampled event times, and committed
marks.  Condition C8 then gives equality of the
states through the common recursion
\(X_j=\Phi(X_{j-1},Z_j)\).  This proves equality of the complete CTMC path laws before the final
completion step in \cref{eq:path-law-equality}. DirectCTMC and SpecCTMC
therefore have the same distribution immediately before completion.
Both methods then apply the same deterministic rule to fill positions
that remain masked at the cutoff, so their final outputs also have the
same distribution.
\end{proof}

\subsection{Uniformization for state-dependent exit rates}

For a generator with state-dependent exit rates, suppose a measurable
candidate intensity \(\Lambda_t\) dominates the exit rate of every
reachable state and has finite integral over the simulated interval.
A nonhomogeneous Poisson candidate clock of intensity \(\Lambda_t\) is
independent of the mark history and satisfies Condition C2.  At a
candidate time, define
\begin{equation}
  P_{t,x}(z)=\frac{q_t(x,z)}{\Lambda_t}
\end{equation}
for each natural mark and add a virtual self mark
\(z_{\mathrm{self}}\) with
\begin{equation}
  P_{t,x}(z_{\mathrm{self}})
  =1-\frac{\lambda_t(x)}{\Lambda_t},
  \qquad
  \Phi(x,z_{\mathrm{self}})=x.
\end{equation}
The proof of \cref{thm:marked-path} applies to these candidate marks.
Removing the virtual self marks recovers the natural CTMC path by
uniformization~\citep{chen2024uniformization}.  Candidate marks need not
remove a mask; validity follows from the independent candidate clock,
not from \cref{lem:retained-clock}.  Conversely, the absorbing-mask
clock calculation does not justify sampling natural event times in advance for a
generator whose future exit rates depend on corrected mark identities.

\section{Additional Theory: Parallel Complexity}
\label{app:additional-theory}

This section proves a complexity bound for a specified adaptive-depth
version of \textsc{SpecCTMC}. The exactness result in \cref{thm:marked-path} does not require the influence
assumption below.  The influence assumption is used only for the cost
bounds.

\subsection{Total unmasking influence}

The prompt and any image conditioning are fixed.  For a partial response
\(x\), let \(A(x)\) be its masked positions and let
\(p_d(\cdot\mid x)\) be the normalized distribution over replacement
tokens at masked position \(d\).

\begin{assumption}[Uniform total unmasking influence]
\label{ass:reveal-influence}
The predictor is time-independent.  There is a constant
\(\Gamma\geq0\) such that every target-valid one-token unmasking event
\(x\to x'\) between reachable states satisfies
\begin{equation}
  \sum_{d\in A(x')}
  \operatorname{TV}\!\left(
    p_d(\cdot\mid x'),p_d(\cdot\mid x)
  \right)
  \leq\Gamma.
\label{eq:reveal-influence}
\end{equation}
Every target row also satisfies Condition C6 of
\cref{app:exactness-assumptions}.
\end{assumption}

The sum in \cref{eq:reveal-influence} permits a large change at a few
positions.  It bounds the total change across predictions at positions
that remain masked after one unmasking event.

\begin{lemma}[Age of the cached target tensor]
\label{lem:cache-age-invariant}
At the beginning of every round, the cached target tensor was evaluated
either at the current committed state or at
an actual ancestor separated from that state by one target-valid unmasking event.
The cached tensor is the last row reached by verification, not a row
from the discarded suffix.
\end{lemma}

\begin{proof}
Proceed by induction over rounds.  The initial target-model evaluation computes probabilities at the
current state, so the claim holds before the first round.  Suppose it
holds at the start of a round.  If the complete draft is accepted, the
last reached target row is evaluated at the state immediately before the
final committed event. Committing that event moves from this state to the
new current state by one target-valid unmasking event.  If verification rejects
at local index \(i\), the last reached target row is evaluated at the
state immediately before the corrected event.  The residual correction is a valid
mark under Conditions C3 and C4, so committing it again places the new
current state one target-valid unmasking event after the row's input state.
Condition C5 discards every later row. Using the last reached row as the
next round's cache therefore preserves the invariant.
\end{proof}

The executed sampler follows this update: it assigns the target tensor
inside the sequential verification loop and, after a rejection, exits
the loop before any later target row can replace it.  It then copies
that last reached tensor into the next round's cache.  The adaptive-depth implementation selects the next depth from
completed-round information before drawing that round's proposals and
verification variables.  These source properties match
\cref{lem:cache-age-invariant} and Condition C7.

\begin{lemma}[Uniform-position dilution]
\label{lem:uniform-position-dilution}
Consider a reached tentative event with local index \(i\) in a round
that began with \(m\) masked positions.  Suppose the proposal assigns
uniform mass to the positions that are currently masked and holds
fixed the cached tensor described in \cref{lem:cache-age-invariant}.  Then the
target and proposal distributions on the joint position-token mark
satisfy
\begin{equation}
  \operatorname{TV}(P_i,D_i)
  \leq\frac{i\Gamma}{m-i+1}.
\label{eq:dilution-bound}
\end{equation}
If the tensor was evaluated at the current state at the beginning of
the round, the numerator improves to \((i-1)\Gamma\).
\end{lemma}

\begin{proof}
Let \(x\) be the state reached before the current update and \(A=A(x)\).  Let \(y_0\) be the
state at which the cached tensor was evaluated.  By
\cref{lem:cache-age-invariant} and because reaching index \(i\) requires
the preceding draft marks to have been accepted, there is a chain of
target-valid unmasking events
\begin{equation}
  y_0,y_1,\ldots,y_\ell=x,
  \qquad \ell\leq i.
\end{equation}
If the cache is current at round entry, \(\ell\leq i-1\).

The target and proposal have the same uniform marginal on \(A\).
Consequently,
\begin{equation}
\begin{aligned}
  \operatorname{TV}(P_i,D_i)
  &=\frac1{|A|}\sum_{d\in A}
    \operatorname{TV}\!\left(
      p_d(\cdot\mid y_\ell),p_d(\cdot\mid y_0)
    \right)\\
  &\leq\frac1{|A|}
    \sum_{k=1}^{\ell}\sum_{d\in A}
    \operatorname{TV}\!\left(
      p_d(\cdot\mid y_k),p_d(\cdot\mid y_{k-1})
    \right)\\
  &\leq\frac{\ell\Gamma}{|A|}
  \leq\frac{i\Gamma}{m-i+1}.
\end{aligned}
\label{eq:dilution-telescope}
\end{equation}
The first equality uses the common uniform position marginal.  The first
inequality is the triangle inequality along the unmasking-event chain. At event
\(k\), the final set \(A\) of masked positions is a subset of the
positions that remain masked after that event, so \cref{ass:reveal-influence} bounds the restricted
inner sum by \(\Gamma\).  Finally, a reached prefix contains only
accepted target-valid marks almost surely, and \(|A|=m-i+1\).  Replacing
\(\ell\leq i\) by \(\ell\leq i-1\) proves the stated improvement.
Time independence in \cref{ass:reveal-influence} is needed here because
\cref{eq:dilution-telescope} contains no additional time-change term.
\end{proof}

\subsection{Adaptive-depth complexity}

Let \(N_0\) be the initial number of masked response positions, let
\(b_{\max}\) be a positive integer, and put
\(\bar\Gamma=\max\{1,\Gamma\}\).  Before drawing the proposal and
verification variables for a round with \(m\) masked positions, choose
\begin{equation}
  K(m)=\max\left\{
    1,
    \min\left\{
      b_{\max},
      \left\lfloor\frac14\sqrt{m/\bar\Gamma}\right\rfloor
    \right\}
  \right\}.
\label{eq:adaptive-depth}
\end{equation}

\begin{theorem}[Influence-based parallel complexity]
\label{thm:influence-complexity}
Under \cref{ass:reveal-influence} and Conditions C1 to C8 of
\cref{app:exactness-assumptions}, use the proposal based on the cached target tensor in
\cref{lem:uniform-position-dilution}, sequential maximal-coupling
verification, correction from the full joint position-token residual,
and the depth in \cref{eq:adaptive-depth}.  For the variant that
completes all \(N_0\) natural unmasking events, the sampler preserves the
complete marked CTMC law.  Excluding initialization, deterministic
completion at the cutoff, and padded physical rows, its logical
target-row count \(W\) and target-batch count \(H\) satisfy
\begin{equation}
  \E[W]\leq\frac87N_0,
\label{eq:influence-work}
\end{equation}
\begin{equation}
  \E[H]\leq\frac87\left(
    \frac{N_0}{b_{\max}}+16\sqrt{N_0\bar\Gamma}
  \right).
\label{eq:influence-rounds}
\end{equation}
The acceptance decisions and round lengths may be dependent and
history-dependent.
\end{theorem}

\begin{proof}
Exactness follows from \cref{thm:marked-path}.  The depth is selected
from pre-round information, every reached row is the declared target
kernel, and correction uses the full joint residual before the suffix is
discarded.

Index rounds by \(r=0,\ldots,R-1\).  Let \(m_r\) be the masks remaining
before round \(r\), \(K_r=K(m_r)\), and \(C_r\) the events committed in
that round, including a corrected event after rejection.  Define the
charged logical rows
\begin{equation}
  \widetilde W=\sum_{r<R}K_r.
\end{equation}
Each round before termination satisfies
\begin{equation}
\begin{aligned}
  1&\leq C_r\leq K_r\leq\min\{b_{\max},m_r\},\\
  R&\leq N_0,
  &\sum_{r<R}C_r&=N_0.
\end{aligned}
\label{eq:round-accounting-complexity}
\end{equation}
The actual logical rows satisfy \(W\leq\widetilde W\), and each round
uses at most one target batch, so \(H\leq R\).

\paragraph{First-rejection probability.}
Fix the pre-round sigma-field \(\mathcal F_r\), and write \(m=m_r\) and
\(K=K_r\).  If \(K\geq2\), then the definition of \(K\) implies
\begin{equation}
  m\geq64\bar\Gamma,
  \qquad K\leq m/2,
  \qquad K+1\leq3K/2.
\label{eq:depth-consequences}
\end{equation}
At a reached index \(i\), the conditional rejection probability equals
\(\operatorname{TV}(P_i,D_i)\) by
\cref{lem:conditional-maximal-coupling}.  The events of first rejection
at different indices are disjoint.  Summing their probabilities and
using \cref{eq:dilution-bound} gives
\begin{equation}
\begin{aligned}
  \Pr(\text{round rejection}\mid\mathcal F_r)
  &\leq\sum_{i=1}^{K}\frac{i\Gamma}{m-i+1}\\
  &\leq\frac{2\Gamma}{m}\sum_{i=1}^{K}i\\
  &=\frac{\Gamma K(K+1)}{m}\\
  &\leq\frac{3\Gamma K^2}{2m}
  \leq\frac3{32}.
\end{aligned}
\label{eq:round-rejection}
\end{equation}
No independence among acceptance decisions is used in this calculation.

\paragraph{Conditional progress.}
If the round has no rejection, it commits \(K\) events.  A rejected
round still commits its corrected event.  Hence
\begin{equation}
\begin{aligned}
  \E[C_r\mid\mathcal F_r]
  &\geq K\Pr(\text{no rejection}\mid\mathcal F_r)\\
  &\geq\frac{29}{32}K
  \geq\frac78K.
\end{aligned}
\label{eq:round-progress}
\end{equation}
If \(K_r=1\), acceptance or correction commits exactly one event, so
\(C_r=1\) almost surely and \cref{eq:round-progress} again holds.

\paragraph{Logical target rows.}
Extend \(C_r\) and \(K_r\) by zero after termination.  Since
\(R\leq N_0\), all sums below have the deterministic upper limit
\(N_0\).  The tower property and \cref{eq:round-progress} yield
\begin{equation}
\begin{aligned}
  N_0
  &=\sum_{r=0}^{N_0-1}\E[C_r]\\
  &=\sum_{r=0}^{N_0-1}
    \E\!\left[\E(C_r\mid\mathcal F_r)\right]\\
  &\geq\frac78\sum_{r=0}^{N_0-1}\E[K_r]
  =\frac78\E[\widetilde W].
\end{aligned}
\label{eq:finite-work-sum}
\end{equation}
Since \(W\leq\widetilde W\), this proves
\cref{eq:influence-work}.

\paragraph{Target batches.}
Define
\begin{equation}
  F(0)=0,
  \qquad
  F(m)=\frac{m}{b_{\max}}+16\sqrt{\bar\Gamma m}
  \quad(m\geq1).
\end{equation}
For \(1\leq C\leq m\), rationalizing the square-root difference gives
\begin{equation}
\begin{aligned}
  F(m)-F(m-C)
  &=\frac{C}{b_{\max}}
    +\frac{16\sqrt{\bar\Gamma}\,C}
           {\sqrt m+\sqrt{m-C}}\\
  &\geq\left(
    \frac1{b_{\max}}+8\sqrt{\frac{\bar\Gamma}{m}}
  \right)C.
\end{aligned}
\label{eq:potential-decrease}
\end{equation}
We now bound the expected decrease on every round before termination.

If capacity binds, \(K=b_{\max}\).  The linear term in
\cref{eq:potential-decrease} and \cref{eq:round-progress} give an
expected decrease of at least \(7/8\).

In the square-root regime, \(K\geq2\) and \(K<b_{\max}\).  With
\(a=\frac14\sqrt{m/\bar\Gamma}\), we have \(a\geq2\) and therefore
\begin{equation}
  K=\lfloor a\rfloor\geq a/2
  =\frac18\sqrt{m/\bar\Gamma}.
\end{equation}
The square-root term in \cref{eq:potential-decrease}, together with
\cref{eq:round-progress}, again gives expected decrease at least
\(7/8\).

In the remaining regime, \(K=1<b_{\max}\).  Then
\(m<64\bar\Gamma\) and \(C=1\) almost surely.  The square-root term in
\cref{eq:potential-decrease} is strictly greater than one, so the
potential decrease is also greater than one.

Set \(m_r=0\) after termination.  Pathwise telescoping over the
nonrandom upper limit \(N_0\) and the three cases above give
\begin{equation}
\begin{aligned}
  F(N_0)
  &=\E\sum_{r=0}^{N_0-1}
    [F(m_r)-F(m_{r+1})]\\
  &\geq\frac78\sum_{r=0}^{N_0-1}\Pr(r<R)
  =\frac78\E[R].
\end{aligned}
\label{eq:finite-potential-sum}
\end{equation}
Thus \(\E[H]\leq\E[R]\leq(8/7)F(N_0)\), which is
\cref{eq:influence-rounds}.  The deterministic finite sums avoid an
optional-stopping argument.
\end{proof}

The depth rule in \cref{eq:adaptive-depth} is the variant analyzed by
\cref{thm:influence-complexity}.  A different adaptive policy requires a
separate complexity argument.

\paragraph{Uniform proposal mixture.}
Let \(U_A\) be the uniform proposal over the valid position-token
marks at the current state, and define
\begin{equation}
  D_i^\eta=(1-\eta)D_i+\eta U_A.
\end{equation}
Convexity of total variation gives
\begin{equation}
\begin{aligned}
  \operatorname{TV}(P_i,D_i^\eta)
  &\leq(1-\eta)\operatorname{TV}(P_i,D_i)
       +\eta\operatorname{TV}(P_i,U_A)\\
  &\leq\operatorname{TV}(P_i,D_i)+\eta.
\end{aligned}
\label{eq:mixture-tv}
\end{equation}
The first-rejection calculation in \cref{eq:round-rejection} therefore
increases by at most \(K\eta\).  If
\(b_{\max}\eta\leq1/32\), its upper bound is
\(3/32+K\eta\leq1/8\).  The conditional progress bound remains
\(\E[C_r\mid\mathcal F_r]\geq(7/8)K_r\), so the constants in
\cref{eq:influence-work,eq:influence-rounds} are unchanged.

\paragraph{Padding and cutoff.}
If every batch is padded to \(b_{\max}\) physical rows, then
\(W_{\mathrm{physical}}\leq b_{\max}H\), and
\cref{eq:influence-rounds} gives
\begin{equation}
  \E[W_{\mathrm{physical}}]
  \leq\frac87N_0
  +\frac{128}{7}b_{\max}\sqrt{N_0\bar\Gamma}.
\end{equation}
Theorem 2 completes all \(N_0\) natural unmasking events.  A simulation stopped
at a positive cutoff has a random number \(N\) of event times at or above
the cutoff; substituting that number into the theorem requires a separate
argument. Target-model evaluations used for initialization and
deterministic completion are excluded from Theorem 2 and are added
separately in \cref{app:proofs}.

The theorem bounds logical target rows and sequential target batches,
not wall-clock latency.  It preserves the chosen reference CTMC under
Conditions C1 to C8; it does not identify that CTMC with the data
distribution.

\subsection{A quality and complexity separation}

For the perfectly dependent pair target \((a=1)\) of
\cref{lem:parallel-dependence}, one unmasking event changes only the
binary prediction of its partner when that partner remains masked, by
total variation \(1/2\).  Thus
\(\Gamma=1/2\) and \(\bar\Gamma=1\).  With batch capacity proportional
to \(\sqrt{N_0}\), \cref{thm:influence-complexity} gives
\begin{equation}
  \E[H]=O(\sqrt{N_0}),
  \qquad
  \E[W]=O(N_0),
\end{equation}
while preserving every pairwise equality constraint.  By contrast, the
preselected independent schedules covered by
\cref{lem:parallel-dependence} require \(\Omega(N_0)\) target-model
evaluations at denoising steps with at least one assigned position to
maintain any fixed high validity level.
This comparison is not a lower bound against adaptive or
dependency-aware parallel samplers.

\section{Extended Experimental Results}
\label{app:experiments}

\subsection{Evaluation protocol}
\label{app:protocol}

\paragraph{Models, tasks, and prompts.}
The primary study uses the public LLaDA-8B-Instruct~\citep{nie2025llada} and Dream-v0-Instruct-7B~\citep{ye2025dream} checkpoints. GSM8K uses five demonstrations, MATH four demonstrations from the \texttt{minerva\_math} configuration, MBPP three demonstrations, and HumanEval and IFEval none; the complete sequence length therefore differs widely across benchmarks at the same generated length. Every model, benchmark, and generated-length condition uses the same fixed evaluation set and five sampling seeds per method. HumanEval and MBPP use the Fast-dLLM code extraction followed by the official benchmark tests in an isolated process with time and memory limits; parse failures, test failures, and timeouts count as incorrect. LLaDA-V uses fixed image conditioning on MathVerse and MathVista with 4,096 validated generations; its task scores are not reported because the required hosted evaluator was unavailable.

\paragraph{Depth selection and timing.}
For every model, benchmark, and generated length, we compare depths 2 through 8 using warmed timing on a fixed selection set. The depth that minimizes \method total generation time is used for the accuracy, NFE, and timing results of that condition. DirectCTMC evaluates one target state per call, while \method evaluates a batch equal to its selected depth. Primary timing uses one exclusively held NVIDIA B200 with balanced method order; model loading and warmup are outside the measured interval, and total generation time includes drafting, target evaluation, verification, residual correction, host-device synchronization, and deterministic completion at the cutoff. The L40S measurements in \cref{fig:depth-hardware,fig:forward-cost-hardware} use one NVIDIA L40S with 46 GiB of memory.

\paragraph{Pairing and intervals.}
Paired comparisons share the precomputed event-time sequence across methods. Target marks, proposal marks, acceptance tests, and residual samples use separate recorded random streams, so DirectCTMC and \method need not produce identical answers. Accuracy and NFE intervals resample questions while keeping the five seeds nested within each question. Timing takes each method's median over its three repetitions within a question, forms the ratio, and bootstraps over questions. A condition is classified as faster or slower only when the paired 95\% interval of \(T_D/T_S\) lies entirely above or below one; an interval containing one is parity. Accuracy difference always means \method minus DirectCTMC.

\subsection{Token choice and unmasking order}
\label{app:token-choice-reveal-order}

Lowering token temperature improves the accuracy of an individual
stochastic answer, but it does not turn DirectCTMC or SpecCTMC into Greedy denoising. Each masked-diffusion update makes two decisions. It chooses which masked position to unmask and then chooses which token to place at that position. The experiment below tests all four combinations of uniform or confidence-based position choice and sampled or highest-probability token choice.

\begin{table}[t]
\centering
\small
\caption{Position choice and token choice for LLaDA-8B-Instruct on GSM8K with 256 generated tokens. Accuracy is reported in percent.}
\label{tab:position-token-choice}
\resizebox{\columnwidth}{!}{%
\begin{tabular}{llr}
\toprule
Position choice & Token choice & Accuracy \\
\midrule
Uniform random & Sampled & 26.56\% \\
Highest confidence & Sampled & 62.50\% \\
Uniform random & Highest probability & 26.56\% \\
Highest confidence & Highest probability & 79.69\% \\
\bottomrule
\end{tabular}%
}
\end{table}

With uniform random unmasking order, replacing token sampling with the highest-probability token leaves accuracy at 26.56 percent. Changing only the unmasking order to highest-confidence selection raises accuracy from 26.56 to 62.50 percent. Combining highest-confidence position selection with the highest-probability token gives 79.69 percent.

In this experiment, confidence-based unmasking order improves accuracy by 35.94 percentage points when tokens are sampled and by 53.13 points when the highest-probability token is used. The highest-probability token adds no measured improvement under uniform random unmasking order, but it adds 17.19 points under confidence-based unmasking order. Thus, the two choices interact, and the unmasking order accounts for most of the measured gap between stochastic CTMC sampling and Greedy denoising. Lower token temperature changes the token distribution while preserving the CTMC's random unmasking order. Greedy denoising instead fixes both the position order and token choice. Random unmasking order lowers the measured single-answer accuracy in this comparison and permits different trajectories across repeated attempts. \Cref{sec:experiments} reports the corresponding Pass@\(K\) results.

\subsection{Wallclock vs NFE reduction}
\label{app:nfe-wallclock}

Let \(N_D\) and \(N_S\) denote the sequential target-model evaluations
made by DirectCTMC and SpecCTMC, and let \(T_D\) and \(T_S\) denote
their total generation times. Their measured speed ratio can be
written as
\begin{equation}
\frac{T_D}{T_S}
=
\frac{N_D}{N_S}
\frac{T_D/N_D}{T_S/N_S}.
\label{eq:nfe-wallclock-decomposition}
\end{equation}
The first factor is the reduction in sequential evaluations. The second
compares the average cost of those evaluations, including candidate
batching and the remaining sampling work. This is the ratio form of the
standard speculative-decoding cost decomposition in
\citet{ho2026specfold}.

The forward-only experiment in
\cref{fig:forward-cost-hardware}a,b verifies that candidate states enter
one model call rather than a Python loop. For three states with complete sequence length \(n\) and hidden width
\(h\), the hidden tensor has shape \([3,n,h]\), and linear layers process
it as \(3n\) token vectors in one matrix multiplication. This improves
wall-clock time only while a one-state evaluation leaves GPU capacity
unused. As sequence length grows, one state already occupies the
available compute and memory bandwidth, so evaluating three states
together performs nearly three times the work and approaches the cost of
three sequential evaluations. Consequently, reducing sequential
target-model evaluations does not produce a comparable wall-clock
reduction for long sequences. At sequence length 911 on L40S, one three-state LLaDA call takes
0.2608 seconds, compared with 0.2547 seconds for three one-state calls.
The same states on B200 take 0.0559 and 0.0640 seconds, respectively.
The four curves extend this comparison over 17 sequence lengths. At
length 400, three states cost 1.33 single-state calls for LLaDA and 1.38
for Dream on B200. At length 3,100, the costs rise to 2.88 and 3.01.

The complete-generation comparison in
\cref{fig:forward-cost-hardware}c uses Dream-7B-Instruct with 256
generated tokens. HumanEval has much shorter inputs than five-shot
GSM8K. On B200, every HumanEval depth is faster than DirectCTMC; on
L40S, depths 2 to 5 are faster and depths 6 to 8 are at parity. For
GSM8K, B200 depths 2 to 4 are at parity and larger depths are slower.
All measured L40S depths are slower, and depth 8 exceeds its memory
capacity. Each curve compares methods on the same five questions and
GPU. The bands show 95 percent intervals obtained by resampling those
questions.

The separate experiment on an exclusively held NVIDIA B200 gives the 2,048-token limiting case. At depth 4,
SpecCTMC reduces sequential target-model evaluations by a factor of
3.9726 with acceptance 0.9950. One depth-4 target call costs 4.0246
one-state calls, so the reduction in sequential evaluations is offset by
the cost of processing four complete states. The measured DirectCTMC to
SpecCTMC time ratio is 0.9871, with a 95 percent interval from 0.9836 to
0.9892. Computing vocabulary projections only at generated positions
preserves complete trajectories while improving total generation time by factors
from 1.062 to 1.136 and reducing allocated memory by 6,742 to 13,530
MiB, showing that implementation choices can change the second factor in
\cref{eq:nfe-wallclock-decomposition}. With LLaDA-8B-Instruct on GSM8K at 2,048 generated tokens, DirectCTMC completes on the L40S, while depth-four SpecCTMC exceeds device memory because one target-model evaluation processes four complete sequence states together.

The role of cache reuse follows the same decomposition. In the Self
Speculative Decoding control~\citep{gao2025ssd}, the cache-disabled
method uses 2.67 times fewer sequential calls but is 1.76 times slower.
With its approximate cache, it uses 2.29 times fewer calls and is 1.40
times faster. The cache lowers the work performed for each candidate
state, which lets the reduction in sequential evaluations reduce elapsed
time. This agrees with \citet{chen2024magicdec}, who show that
key/value-cache reuse can restore speculative-decoding gains when
long-context autoregressive decoding is limited by memory traffic.

\subsection{Deterministic Decoding References}
\label{app:deterministic-references}

\Cref{tab:deterministic-matched} reports the fully matched
comparison used to define the deterministic reference in
\cref{fig:stochastic-coverage}.
Greedy denoising selects the highest-confidence position and
its highest-probability token.
The Fast-dLLM threshold and factor configurations use
confidence-guided parallel decoding~\citep{wu2025fastdllm}.
All three configurations are deterministic under the tested
settings.
Repeating an identical prompt therefore produces the same
answer, and empirical Pass@\(K\) remains equal to single-answer
accuracy.
The strongest matched result is 81.25\% from Fast-dLLM factor 1.0.

\begin{table*}[t]
\centering
\small
\setlength{\tabcolsep}{4pt}

\caption{
Matched deterministic configurations on LLaDA-8B-Instruct and
five-shot GSM8K with 256 generated tokens.
All rows use the same checkpoint revision, prompts, scorer,
and timing procedure.
Scores and costs are our measured results.
These configurations are deterministic under the tested
settings, so repeated identical runs have the same Pass@\(K\)
as their single-answer accuracy.
}
\label{tab:deterministic-matched}

\begin{tabular}{llrrr}
\toprule
Method &
Selection rule &
Accuracy (\%) &
Mean NFEs &
Median time (s) \\
\midrule

Greedy denoising &
\shortstack[l]{
  Highest-confidence position;\\
  highest-probability token
} &
79.69 & 256.00 & 34.78 \\

Fast-dLLM, threshold 0.9 &
Confidence threshold \(>0.9\) &
78.13 & 80.11 & 10.52 \\

Fast-dLLM, factor 1.0 &
Confidence-profile factor rule &
\textbf{81.25} & 63.06 & 8.13 \\

\bottomrule
\end{tabular}
\end{table*}

\Cref{tab:deterministic-other-implementations} reports additional
greedy-derived speculative implementations, including PSD,
trajectory-level speculative decoding, and
Spiffy~\citep{sun2026psd,pan2026trajectory,agrawal2026spiffy}.
Each implementation is reported beside the Greedy reference
measured with that implementation's evaluation setup.
The reference line in the main figure comes from the fully matched
comparison in \cref{tab:deterministic-matched}.

\begin{table*}[t]
\centering
\small
\setlength{\tabcolsep}{4pt}

\caption{
Additional greedy-derived speculative implementations.
Each row is reported with its own measured Greedy reference because
implementation, input shape, or cache definition differs.
The final column reports the within-implementation ratio of Greedy time
to method time; a value above one indicates acceleration relative to
that reference. Absolute latency is not pooled across these
implementations. Each method is named according to its evaluated
configuration, and all numerical results are from our runs.
}
\label{tab:deterministic-other-implementations}

\begin{tabular}{p{0.43\textwidth}rrrr}
\toprule
Method/configuration &
\shortstack{Own Greedy\\accuracy (\%)} &
\shortstack{Method\\accuracy (\%)} &
\shortstack{Mean\\NFEs} &
\shortstack{Greedy time /\\method time} \\
\midrule

Parallel Speculative Decoding, confidence 0.9 &
79.69 & 76.56 & 60.39 & 1.684 \\

Trajectory-Level, single-block exact &
73.44 & 73.44 & 160.17 & 1.544 \\

Trajectory-Level, practical cross-block &
73.44 & 76.56 & 203.44 & 1.016 \\

Self Cascaded Speculative Decoding, no cache &
82.81 & 79.69 & 81.59 & 0.599 \\

Self Cascaded Speculative Decoding, approximate cache &
73.44 & 78.13 & 100.98 & 1.268 \\

Spiffy, independent-state implementation &
79.69 & 78.13 & 127.19 & 0.261 \\

Spiffy, practical tree implementation &
79.69 & 75.00 & 128.66 & 0.333 \\

\bottomrule
\end{tabular}
\end{table*}

\subsection{Approximate Solver Details}
\label{app:approximate-solvers}

This subsection defines the two finite-step continuous-time Markov chain
solvers used in \cref{sec:approximate-samplers}.

Let \(X_t\) be the partially masked sequence at diffusion time \(t\). For masked position \(i\) and candidate token \(v\), let \(q_t(i,v\mid x)\) be the instantaneous transition rate at state \(x\). For time interval \(I_n\) of width
\begin{equation}
h_n=|t_{n+1}-t_n|,
\qquad
\kappa=\max_n h_n,
\label{eq:solver-step-size}
\end{equation}
define the exact integrated intensity
\begin{equation}
\Lambda_{n,i,v}
=
\int_{I_n}q_t(i,v\mid X_t)\,dt.
\label{eq:integrated-intensity}
\end{equation}

\paragraph{Tau leaping.}
Tau leaping freezes the rate at the beginning of the interval~\citep{campbell2022continuous}:
\begin{equation}
\Lambda_{n,i,v}
\approx
h_n q_{t_n}(i,v\mid X_{t_n}),
\label{eq:tau-leaping}
\end{equation}
and samples
\begin{equation}
K_{n,i,v}
\sim
\operatorname{Poisson}\!\left(
h_n q_{t_n}(i,v\mid X_{t_n})
\right).
\label{eq:tau-poisson}
\end{equation}
Let \(K_{n,i}=\sum_v K_{n,i,v}\). The implementation leaves position \(i\) masked when \(K_{n,i}=0\), unmasks the position with the unique sampled token when \(K_{n,i}=1\), and discards all competing events at that position when \(K_{n,i}>1\). One model evaluation can unmask many different positions. The approximation is
\begin{equation}
q_t(i,v\mid X_t)
\approx
q_{t_n}(i,v\mid X_{t_n})
\qquad\text{for }t\in I_n.
\label{eq:tau-frozen-rate}
\end{equation}
After one position is unmasked, the exact rates at other positions can change, while tau leaping continues to use the rates from the interval start. Shorter intervals reduce rate staleness and conflicts but require more model evaluations.

\paragraph{Second-order theta-trapezoidal solver.}
The trapezoidal method uses a predictor and correction to estimate how rates change within the interval~\citep{ren2025fast}. The experiments use \(\theta=1/2\). Define
\begin{equation}
q_n(i,v)=q_{t_n}(i,v\mid X_{t_n}).
\end{equation}
The predictor samples
\begin{equation}
K^{\mathrm{pred}}_{n,i,v}
\sim
\operatorname{Poisson}\!\left(\frac{h_n}{2}q_n(i,v)\right),
\label{eq:trap-predictor}
\end{equation}
which gives an intermediate state \(X^*\). The second model evaluation gives
\begin{equation}
q_n^*(i,v)
=
q_{(t_n+t_{n+1})/2}(i,v\mid X^*).
\label{eq:trap-midpoint-rate}
\end{equation}
Let \(q_n^{\mathrm{aligned}}\) equal the initial rate at positions left unchanged by the predictor and zero at positions already unmasked. The implemented correction rate is
\begin{equation}
q_n^{\mathrm{corr}}
=
\left[
\frac{1}{2\theta}q_n^*
-
\frac{(1-\theta)^2+\theta^2}{2\theta}
q_n^{\mathrm{aligned}}
\right]_+,
\label{eq:trap-correction-general}
\end{equation}
where \([a]_+=\max(a,0)\). At \(\theta=1/2\),
\begin{equation}
q_n^{\mathrm{corr}}
=
\left[q_n^*-\frac{1}{2}q_n^{\mathrm{aligned}}\right]_+.
\label{eq:trap-correction-midpoint}
\end{equation}
The corrector samples
\begin{equation}
K^{\mathrm{corr}}_{n,i,v}
\sim
\operatorname{Poisson}\!\left(h_n q_n^{\mathrm{corr}}(i,v)\right).
\label{eq:trap-corrector}
\end{equation}
Thus the method uses two rate evaluations per configured step.

Under the regularity, rate-estimation, and positivity assumptions of \citet{ren2025fast}, the discretization terms over diffusion-time horizon \(T\) scale as
\begin{equation}
\mathcal E_{\mathrm{disc}}^{\mathrm{tau}}=O(\kappa T),
\qquad
\mathcal E_{\mathrm{disc}}^{\mathrm{trap}}=O(\kappa^2T).
\label{eq:solver-error-orders}
\end{equation}
This is a convergence statement as \(\kappa\) approaches zero. It does not imply higher downstream task accuracy at every coarse evaluation budget. For nonzero step size, both methods approximate the DirectCTMC process.


\subsection{Composition with caching and Fast-dLLM-v2 self-drafting}
\label{app:composition}

These experiments test two ways to combine SpecCTMC with other diffusion
language model improvements. Prompt caching changes the cost and the
conditional probabilities of the target evaluator. Fast-dLLM-v2
self-drafting uses the same checkpoint to compute the proposal and the
current target probabilities. In both cases, SpecCTMC verifies each
proposed event against the declared sequential transition probabilities
and samples the positive residual after the first rejection.

\paragraph{Prompt caching.}
The prompt-cache experiment uses LLaDA-8B-Instruct on five-shot GSM8K
with 256 generated tokens. Both cached methods use target batch
size three. Cached SpecCTMC uses speculation depth three. The cache stores
key/value tensors for the fixed prompt, recomputes all generated
positions, and refreshes the prompt representation after 8, 16, or 32
committed unmasking events. Because bidirectional attention allows generated tokens
to change prompt representations, the cache defines an approximate target
process relative to complete recomputation. SpecCTMC preserves the
matching cached sequential process.

At every refresh interval in \cref{tab:prompt-cache-composition}, the 95 percent interval for the accuracy difference between cached SpecCTMC and the matching cached sequential method contains zero. At refresh intervals 8 and 16, the interval for cached sequential accuracy relative to uncached DirectCTMC also contains zero. At interval 32, the cache lowers cached sequential accuracy from 32.8\% to 17.2\%. This quality change is caused by cache reuse, while SpecCTMC preserves the corresponding cached sequential process.


\begin{table}[t]
\centering
\small
\caption{Task quality for prompt-cached sequential CTMC and prompt-cached SpecCTMC. Changes are percentage points with 95\% CI. Cache change compares cached sequential CTMC with uncached DirectCTMC. SpecCTMC change compares the two cached methods.}
\label{tab:prompt-cache-composition}
\resizebox{\columnwidth}{!}{%
\begin{tabular}{rrrrr}
\toprule
Refresh & Cached seq. & Cached Spec & Cache change & SpecCTMC change \\
interval & accuracy & accuracy & from uncached & from cached seq. \\
\midrule
8  & 25.0\% & 35.9\% & \(-7.8\) [\(-18.8\), \(1.6\)] & \(10.9\) [\(-1.6\), \(23.4\)] \\
16 & 26.6\% & 28.1\% & \(-6.3\) [\(-15.6\), \(3.1\)] & \(1.6\) [\(-10.9\), \(14.1\)] \\
32 & 17.2\% & 20.3\% & \(-15.6\) [\(-26.6\), \(-4.7\)] & \(3.1\) [\(-6.3\), \(12.5\)] \\
\bottomrule
\end{tabular}%
}
\vspace{-1em}
\end{table}

\begin{table}[t]
\centering
\small
\caption{Target-model evaluation accounting for the 256-token prompt-cache experiment. Refresh calls apply to both cached methods. The final two columns compare cached sequential CTMC with cached SpecCTMC at the same refresh interval.}
\label{tab:prompt-cache-evaluations}
\resizebox{\columnwidth}{!}{%
\begin{tabular}{rrrrrrr}
\toprule
Refresh & Seq. target & Spec target & Refresh & Spec total & Call & Time ratio \\
interval & calls & calls & calls & calls & reduction & with 95\% interval \\
\midrule
8  & 253.9 & 100.6 & 32.9 & 133.5 & 2.530 & 1.697 [1.692, 1.700] \\
16 & 253.9 & 98.4  & 17.0 & 115.3 & 2.586 & 1.940 [1.936, 1.949] \\
32 & 253.9 & 92.7  & 9.0  & 101.6 & 2.751 & 2.211 [2.201, 2.217] \\
\bottomrule
\end{tabular}%
}
\end{table}

The evaluation comparison in \cref{tab:prompt-cache-evaluations} counts
each cache reconstruction as an additional sequential model evaluation.
It therefore includes work omitted by a target-call-only count. Cached
SpecCTMC is 1.697, 1.940, and 2.211 times faster than cached sequential
CTMC at refresh intervals 8, 16, and 32. Compared with approximately 256
dependent steps for 256-token autoregressive decoding, The totals of
133.5, 115.3, and 101.6 sequential model evaluations correspond to reductions of 1.92,
2.22, and 2.52 times. At intervals 8 and 16, the confidence interval for the cached sequential accuracy difference from uncached DirectCTMC contains zero. This comparison concerns sequential depth. One autoregressive model
evaluation and one bidirectional SpecCTMC target-model evaluation perform
different amounts of computation.


\paragraph{Fast-dLLM-v2 checkpoint transfer.}
The second experiment uses the Fast-dLLM-v2 7B checkpoint associated with
DualDiffusion~\citep{goyal2026dualdiffusion}. It evaluates GSM8K at 32 generated tokens. The event-by-event reference and SpecCTMC both use physical target batch four, and SpecCTMC uses depth four. Predictions from a previous call to the same checkpoint propose future events. SpecCTMC verifies each proposal using probabilities evaluated at its tentative state and corrects the first rejected proposal.

\begin{table}[t]
\centering
\small
\caption{Fast-dLLM-v2 checkpoint transfer. The same checkpoint is used to compute the event-by-event reference, proposals, and current verification probabilities.}
\label{tab:fastdllmv2-checkpoint-transfer}
\resizebox{\columnwidth}{!}{%
\begin{tabular}{lrr}
\toprule
Quantity & Event by event & SpecCTMC \\
\midrule
Sequential target calls & 32.00 & 10.25 \\
Sequential-call reduction & 1.00 & 3.122 \\
Target states evaluated & 32.00 & 38.50 \\
Additional proposal calls & 0.00 & 1.50 \\
Total checkpoint calls & 32.00 & 11.75 \\
Accepted events per round & Not applicable & 2.631 \\
Proposal acceptance & Not applicable & 70.19\% \\
Residual corrections & 0.00 & 5.75 \\
Median total generation time, seconds & 55.632 & 55.696 \\
\bottomrule
\end{tabular}%
}
\end{table}

The experiment reduces sequential target calls by 3.122 times and total
checkpoint evaluations from 32.00 to 11.75 after proposal calls are
included. The elapsed times are nearly equal in this comparison.
This experiment shows that SpecCTMC can use Fast-dLLM-v2 checkpoint probabilities for both proposal and exact verification. It omits the native cache updates and confidence-based multi-token commitment rule.

\section{Related Work}
\label{app:related-work}

\paragraph{Exact speculative sampling.}
\citet{debortoli2025accelerated} extend speculative sampling to continuous diffusion transitions on a fixed denoising schedule. Their method establishes the linear-draft pattern used here: draft future states, evaluate target transitions in parallel, verify a prefix, correct the first rejection, and discard the suffix. AutoSpec exploits exchangeable increments to parallelize continuous denoising diffusion probabilistic models without an auxiliary drafter~\citep{hu2025autospec}. Both operate on continuous fixed-schedule diffusion rather than a discrete marked CTMC whose event times are sampled before its marks.
\citet{soen2026block} develop efficient continuous residual sampling and joint block verification for speculative diffusions. Joint block verification also improves autoregressive speculative decoding~\citep{sun2024block}. The contribution here is the marked event construction for an event-driven discrete process, including a random clock and a joint categorical position/token residual. We identify the absorbing-mask condition under which event times sampled before the marks remain valid after mark correction.

\paragraph{Masked diffusion speculation and acceleration.}
Parallel masked-diffusion decoders select several positions from one model evaluation. Fast-dLLM and Fast-dLLM++ use confidence rules that mitigate dependency violations and can preserve a greedy mode under stated assumptions~\citep{wu2025fastdllm,kasa2026fastdllmpp}. EB-Sampler controls an entropy-based proxy for dependence error, while PUNT tests approximate contextual independence~\citep{benhamu2025ebsampler,azangulov2025punt}. None of these methods reconstructs the complete parallel update as the distribution of its sequential reference. \citet{chen2025schedules} characterize the expected Kullback-Leibler divergence of parallel masked sampling through the unmasking schedule and the target distribution's information curve. Their result establishes the general dependence-loss mechanism. \citet{ringel2026demask} train a pairwise influence predictor and, under a subadditivity assumption, choose blocks whose factorized update lies within a selected total-variation tolerance of the corresponding joint update. \citet{zhang2026sirs} use randomized rejection increments so that practical clipping replaces the full within-block dependence error with a bounded clipping term. Both control the parallel factorization error without universally eliminating it. SpecCTMC instead uses target evaluation and residual correction to preserve its declared transition kernel. Sequential first-hitting sampling (FHS), which DirectCTMC instantiates, exactly simulates the reverse CTMC one event at a time; its parallel variant uses the same first-hitting times while freezing predictions across several unmasking events and is explicitly an approximation~\citep{zheng2025fhs}. \citet{liang2026sharp} prove in total variation that FHS adds no sampling error beyond score-estimation error, while finite-step Euler sampling has a separate discretization contribution. \Cref{tab:method-positioning} compares where parallelism enters and whether each method preserves exact event times, the natural-event CTMC law, and its stated sequential reference.

A method can add no parallel factorization error while preserving a sequential reference other than DirectCTMC. SpecCTMC may use previously computed factorized predictions because they determine only tentative proposals. Sequential target verification and positive-residual correction over the complete position/token mark space remove the proposal bias, so the committed process adds no parallel factorization error to DirectCTMC.

Self Speculative Decoding reproduces its deterministic stepwise reference through hierarchical verification~\citep{gao2025ssd}. Spiffy is lossless when its batched draft probabilities equal the corresponding vanilla probabilities; its practical simple-mask implementation for fully bidirectional LLaDA and Dream is described as near-lossless~\citep{agrawal2026spiffy}. Spiffy uses the target model as its drafter and calibrates its directed draft graphs offline. Parallel Speculative Decoding constructs multi-depth drafts and selects the deepest draft consistent with updated predictions, reporting a speed and quality tradeoff rather than equality with a stochastic CTMC law~\citep{sun2026psd}. Trajectory-Level Speculative Decoding proves exactness for a restricted deterministic top-one setting with fixed cached states and no inter-block speculation, while its practical cross-block method permits trajectory drift~\citep{pan2026trajectory}. Self-Speculative Masked Diffusions trains a hybrid causal and noncausal architecture whose full-network conditionals define a new nonfactorized target~\citep{campbell2026self}. Probability-ratio acceptance and residual resampling remove factorization error relative to that trained target. These methods establish single-model speculative decoding results for finite-step references, while SpecCTMC combines within-trajectory parallel verification with preservation of an unchanged denoiser's natural-event CTMC law.

SpecCTMC does not require the denoiser's per-position predictions across masking contexts to be conditional marginals of a single joint distribution. DirectCTMC and SpecCTMC target the sequential process specified by the state-dependent transition kernel. The exactness theorem therefore requires normalized transition rows but no compatible any-order conditional oracle.

\paragraph{Exact parallel sampling theory.}
Any-Subset Speculative Decoding samples from trained any-subset autoregressive models using their joint-density oracle~\citep{guo2026anysubset}. Parallel Sampling via Autospeculation gives exact endpoint samples in \(\widetilde O(\sqrt n)\) parallel rounds when an oracle returns compatible conditional marginals of one target distribution~\citep{anari2025autospeculation}. \citet{liu2024parallelize} faithfully parallelize general single-site dynamics under influence conditions in a parallel random-access machine model. Under the bounded-unmasking-influence assumption, Theorem~\ref{thm:influence-complexity} bounds full-network target evaluations for the joint position/token proposal based on cached target predictions, with finite batch capacity. It permits normalized state-dependent predictions that need not be conditional marginals of a single joint distribution.

\paragraph{Temporal point processes.}
TPP-SD uses a smaller draft transformer to propose multiple future temporal-point-process events and parallel target verification~\citep{gong2025tppsd}. \citet{bilos2025tpp} exactly sample multiple future events from unchanged parametric temporal point processes through rejection bounds. These works share continuous event times and marked sequences. Our condition and correction differ: absorbing-mask intensity is independent of mark identities after conditioning on event count, so event times sampled before the marks remain valid after categorical position/token correction through maximal coupling.

\paragraph{Exact and approximate discrete diffusion simulation.}
Discrete diffusion has a standard CTMC formulation, commonly simulated through tau-leaping~\citep{campbell2022continuous}. Uniformization can remove time-discretization error by introducing a common candidate clock with virtual self transitions~\citep{chen2024uniformization}. Higher-order Runge-Kutta and trapezoidal solvers improve approximation order~\citep{ren2025fast}. Uniformization provides an exact serial reference, whereas tau-leaping and higher-order solvers provide approximate comparisons. SpecCTMC instead couples a proposed embedded transition to the target transition and batches neural evaluations at states preceding the drafted updates.

\section{Implementation Details}
\label{app:reproducibility}

Our implementation drafts by reusing previously computed target token probabilities, without the need for an auxiliary model. At each tentative state, the proposal selects a masked position uniformly and samples a token from the reused probability vector for that position. For both checkpoints, we remove the mask-token probability before normalizing the remaining token probabilities. Dream applies the checkpoint's one-position logit shift once before normalization.

Unless stated otherwise, we use temperature one, no classifier-free guidance, and cutoff \(\varepsilon=0.01\). The proposal mixes previously computed target probabilities with a uniform distribution over replacement tokens using weight \(10^{-4}\). The evaluation protocol, including depth selection, target-state grouping, timing, pairing, and code scoring, is described in \cref{app:protocol}.

\paragraph{Proposal and randomness.}
An initial target-model evaluation computes probabilities at the current state. Paired comparisons use the same event-time sequence across methods. Target-mark, draft-mark, verification, and residual draws use separate recorded random streams. Method order alternates across matched workloads.

\paragraph{Benchmarks and scoring.}
We follow the same procedure as described in \citep{wu2025fastdllm, kasa2026fastdllmpp}:GSM8K uses numeric exact extraction. MATH uses symbolic answer equivalence. MBPP and HumanEval use restricted execution that rejects unsafe or unsupported candidates and records the reason. IFEval uses the pinned first 300 prompts in source order, giving 456 instruction instances. Prompt-level strict accuracy is the task score. Strict and loose checkers receive identical deterministic prompt-derived random choices. Dream text projection preserves raw sampled identifiers and maps undefined tokenizer identifiers to the Unicode replacement character only in scorer input.

\end{document}